\pdfoutput=1
\documentclass[11pt]{article}
\usepackage[preprint]{acl}
\usepackage[table]{xcolor}
\usepackage{times}
\usepackage{latexsym}
\usepackage{caption}
\usepackage{float}
\usepackage{booktabs}
\usepackage{amsmath}
\usepackage{amssymb}
\usepackage{hyperref}
\usepackage{xurl}
\usepackage[T1]{fontenc}
\usepackage[utf8]{inputenc}
\usepackage{microtype}
\usepackage{inconsolata}
\usepackage{graphicx}
\title{SPPO: Sequence-Level PPO for Long-Horizon Reasoning Tasks}

\author{
Tianyi Wang$^{1,3}$\thanks{Equal Contribution. $^{\dag}$Corresponding Author.},
Yixia Li$^{1*}$,
Long Li$^{2}$,
Yibiao Chen$^{3}$,\\
\textbf{Shaohan Huang}$^{4}$,
\textbf{Yun Chen}$^{5}$,
\textbf{Peng Li}$^{6}$,
\textbf{Yang Liu}$^{6}$,
\textbf{Guanhua Chen}$^{1\dagger}$\\
$^1$Southern University of Science and Technology,  
$^2$INFLY TECH\\
$^3$Beijing University of Posts and Telecommunications,
$^4$Microsoft Research Asia\\
$^5$Shanghai University of Finance and Economics,
$^6$Tsinghua University 
}

\begin{document}
\maketitle
\begin{abstract} Proximal Policy Optimization (PPO) was central to aligning Large Language Models (LLMs) in reasoning tasks with verifiable rewards. However, standard token-level PPO struggles in this setting due to the instability of temporal credit assignment over long Chain-of-Thought (CoT) horizons and the prohibitive memory cost of the value model. While critic-free alternatives like GRPO mitigate these issues, they incur significant computational overhead by requiring multiple samples for baseline estimation, severely limiting training throughput. In this paper, we introduce \textbf{Sequence-Level PPO (SPPO)}, a scalable algorithm that harmonizes the sample efficiency of PPO with the stability of outcome-based updates. SPPO reformulates the reasoning process as a \textbf{Sequence-Level Contextual Bandit} problem, employing a decoupled scalar value function to derive low-variance advantage signals without multi-sampling. Extensive experiments on mathematical benchmarks demonstrate that SPPO significantly surpasses standard PPO and matches the performance of computation-heavy group-based methods, offering a resource-efficient framework for aligning reasoning LLMs.
\end{abstract}

\section{Introduction}
\label{intro}
Large Language Models (LLMs) have significantly advanced in complex reasoning, empowered by long Chain-of-Thought (CoT) prompting \citep{lightman2023letsverifystepstep}. To further align these models with logical correctness, Reinforcement Learning (RL) has proven indispensable, particularly in Reinforcement Learning with Verifiable Rewards (RLVR) tasks like mathematical problem-solving \citep{deepscaler2025}.
Proximal Policy Optimization (PPO, \citet{schulman2017proximalpolicyoptimizationalgorithms}) typically relies on a token-level Critic and Generalized Advantage Estimation (GAE) for credit assignment \citep{guo2025segmentpolicyoptimizationeffective}. However, this framework faces structural incompatibility in long CoT tasks with sparse rewards. The delayed reward forces GAE to propagate signals across thousands of tokens, inducing high bias \citep{yuan2025whatspposcollapselongcot}. Furthermore, the Critic tends to ``overfit'' semantic cues at the sequence tail (See Figure \ref{fig:value_charts}), causing the advantage signal to vanish precisely when needed. Consequently, standard PPO often proves unstable for reasoning tasks \citep{kazemnejad2025vinepporefiningcreditassignment}.

In response, Group Relative Policy Optimization (GRPO, \citet{shao2024deepseekmathpushinglimitsmathematical}) eliminates the learned Critic in favor of group-based statistical baselines. 
We posit that GRPO's success stems from implicitly remodeling reasoning as a \textbf{Sequence-Level Contextual Bandit} problem, treating the entire response as an atomic action to bypass token-level noise. 
However, this approach faces a fundamental \textbf{Bias-Variance Trade-off}: while it removes the high bias inherent in token-level value estimation, the reliance on Monte Carlo outcomes introduces \textbf{high variance} in the gradient signal \citep{wang2025kalmanfilterenhancedgrpo}. 
To mitigate this variance and stabilize training, GRPO incurs a prohibitive computational cost, as it necessitates sampling multiple responses ($N$) per prompt to construct a valid baseline, significantly bottlenecking training throughput \citep{lin2025cppoacceleratingtraininggroup,li2025adaptivegrouppolicyoptimization}.

Despite the empirical success of critic-free methods like GRPO, a critical misconception exists regarding their efficacy; our core contribution is understanding GRPO from a novel perspective: its success stems from implicitly remodeling reasoning as a Sequence-Level Contextual Bandit problem---treating the entire response as an atomic action to bypass token-level noise---rather than a multi-step Markov Decision Process (MDP). We posit that a stable, sequence-level baseline—underpinned by a generalizable scalar value model—is structurally more robust for long-horizon RLVR tasks. Crucially, this explicitly modeled approach not only secures optimization stability but also circumvents the prohibitive computational latency associated with extensive group sampling.

Drawing on these insights, we introduce \textbf{Sequence-Level PPO (SPPO)}, a novel algorithm that resolves the Bias-Variance dilemma in reasoning alignment. SPPO fundamentally reformulates the reasoning process from a token-level Markov Decision Process (MDP) to a \textbf{Sequence-Level Contextual Bandit} problem. In this view, the prompt serves as the static context and the \textit{entire reasoning chain} is treated as a single atomic action. This formulation effectively collapses the time horizon, eliminating the \textbf{high bias} of token-level credit assignment inherent to standard PPO. Simultaneously, SPPO employs a learned scalar value function to curb the \textbf{high variance} of group-relative baselines, thereby achieving optimization stability without the need for multi-sampling ($N>1$).
\url
Crucially, SPPO addresses computational bottlenecks through this resource-efficient architecture. Unlike GRPO, which requires expensive multi-sampling for empirical baselines to reduce variance, SPPO leverages its learned scalar value function to enable high-throughput single-sample updates ($N=1$). Furthermore, we validate a \textbf{Decoupled Critic} strategy—using a lightweight critic (e.g., 1.5B) to align a larger policy (e.g., 7B)—which leverages the reduced complexity of value estimation to cut memory usage by \textbf{12.8\%} (Figure \ref{fig:memory_usage}). Extensive evaluations on AIME24/25, AMC23, MATH, and Minerva demonstrate that SPPO resolves the value collapse of standard PPO and outperforms computation-heavy baselines. Notably, SPPO matches GRPO's peak performance with a \textbf{5.9$\times$ training speedup} and superior convergence, offering a scalable paradigm for sparse-reward reasoning tasks.\footnote{Our code is available at \url{https://github.com/sustech-nlp/SPPO}.}
\section{Background}

\subsection{PPO and Credit Assignment in Reasoning}
\begin{figure}[t]
    \centering
    \includegraphics[width=1.0\linewidth]{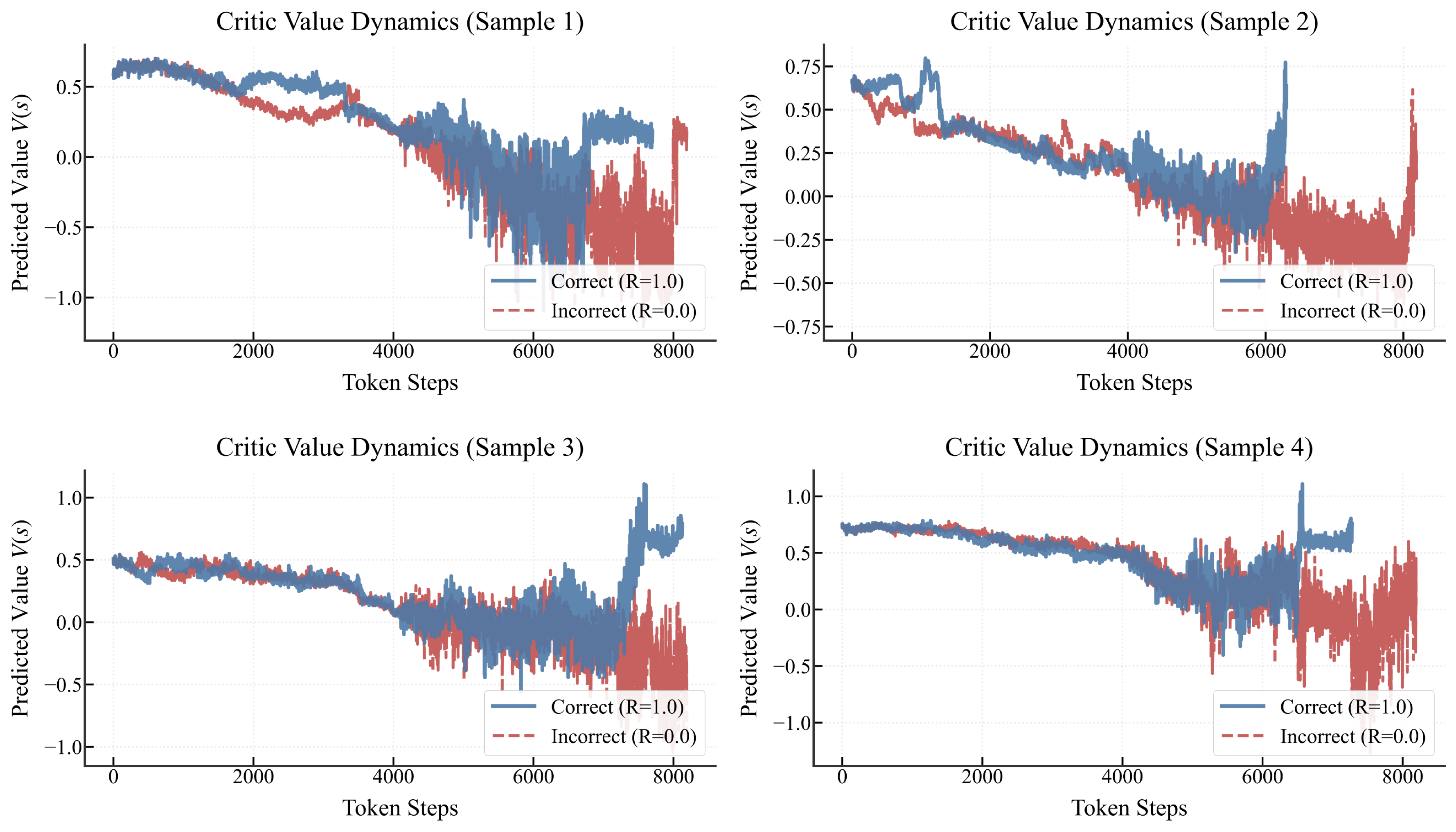} 
    \caption{\textbf{Analysis of the ``Tail Effect''.} 
    We visualize Critic value dynamics $V(s_t)$ to diagnose inefficiencies. Blue and red lines denote correct and incorrect trajectories, respectively. 
    The Critic discriminates only near the sequence tail. For correct paths, $V(s_t)$ rises late, causing $\hat{A}_t$ to vanish; for incorrect ones, it fails to penalize intermediate steps. This indicates credit assignment based on token position rather than semantic contribution. The Critic was trained under 8192 context window. Additional randomly sampled visualizations are provided in Appendix \ref{app:value_vis}.}
    \label{fig:value_charts}
\end{figure}
PPO optimizes the policy by maximizing a clipped surrogate objective $J_{\text{PPO}}(\theta)$:
\begin{align*}
\label{eq:ppo_objective}
J_{\text{PPO}}(\theta) &= \mathbb{E}_t \Big[ \min \big( r_t(\theta) \hat{A}_t, \notag \\
&\quad \text{clip}(r_t(\theta), 1-\epsilon, 1+\epsilon) \hat{A}_t \big) \Big]
\end{align*}

\begin{figure}[b!]
  \centering
  \includegraphics[width=1.0\columnwidth]{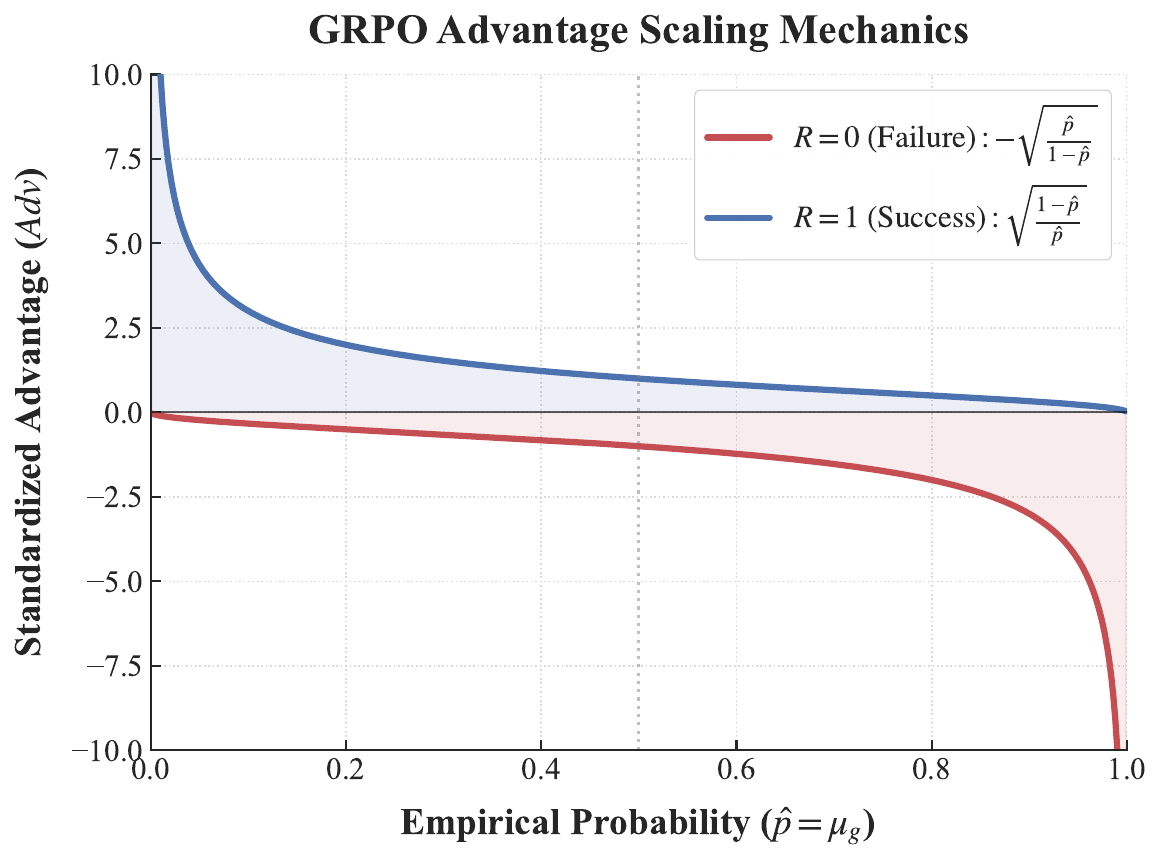}
  \caption{\textbf{Visualization of the GRPO Advantage Function.} derived under the Bernoulli assumption (see Appendix \ref{app:grpo_derivation}). The plot illustrates how GRPO implicitly models the reasoning task as a \textbf{Contextual Bandit}: instead of a static reward, the advantage is dynamically scaled based on the prompt's estimated difficulty $\hat{p}(s_p)$, contrasting success (Blue) against failure (Red).}
  \label{fig:grpo_adv}
\end{figure}

\begin{figure*}[t]
    \centering
    \includegraphics[width=1.0\textwidth]{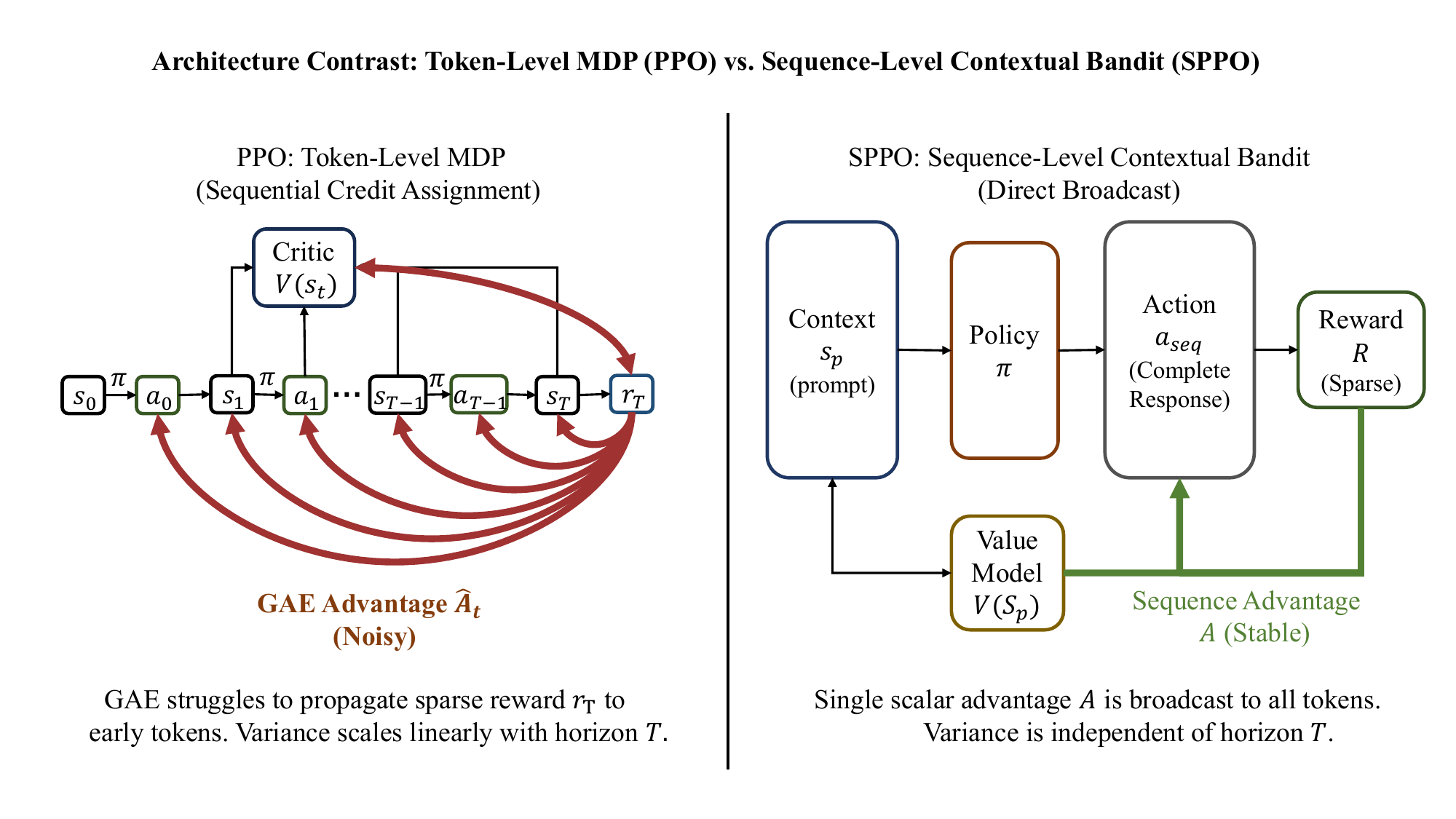}
    \caption{\textbf{Overview of SPPO.} Motivated by the implicit bandit behavior of GRPO, SPPO explicitly reformulates reasoning as a Sequence-Level Contextual Bandit, utilizing a scalar value function $V(s_p)$.}
    \label{fig:main}
\end{figure*}

where $r_t(\theta)$ is the probability ratio and $\hat{A}_t$ is the advantage. Typically, $\hat{A}_t$ is computed via Generalized Advantage Estimation (GAE) using a learned token-level Critic $V(s_t)$. The TD error is defined as $\delta_t = r_t + \gamma V(s_{t+1}) - V(s_t)$, and the advantage is the discounted sum of errors:
\begin{equation*}
    \hat{A}_t^{\text{GAE}} = \sum_{l=0}^{T-t-1} (\gamma\lambda)^l \delta_{t+l}
\end{equation*}
In sparse-reward reasoning tasks (e.g., CoT), it is standard to set $\gamma=\lambda=1$ to propagate terminal rewards. Under this setting, GAE simplifies to the difference between the Monte Carlo return $G_t$ and the value estimate:
\begin{equation*}
    \hat{A}_t^{\text{GAE}} = G_t - V(s_t)
\end{equation*}

However, this mechanism is unstable in long-horizon tasks (Figure \ref{fig:value_charts}). As generation approaches the answer, the Critic $V(s_t)$ often ``overfits'' semantic cues. For correct trajectories, $V(s_t)$ converges to the reward early, causing $\hat{A}_t$ to vanish; for incorrect ones, it underestimates significantly. This ``tail effect'' bases credit on position rather than contribution, undermining optimization.

\subsection{Optimization Mechanics of GRPO}
\label{subsec:grpo_optim}

While standard PPO operates within a token-level MDP, GRPO implicitly shifts the optimization paradigm. It eliminates the step-wise Critic by sampling $N$ outputs per prompt and computing the advantage via group normalization:
\begin{equation*}
    Adv(s_p,a) = \frac{R - \mu_g}{\sigma_g}.
\end{equation*}
Modeling the sampling process as Bernoulli trials (derivation provided in Appendix \ref{app:grpo_derivation}), the advantage function simplifies to:
\begin{equation*}
Adv(s_p,a) = 
\begin{cases}
\sqrt{\frac{1-\hat{p}(s_p)}{\hat{p}(s_p)}} & \text{if } R = 1 \\
-\sqrt{\frac{\hat{p}(s_p)}{1-\hat{p}(s_p)}} & \text{if } R = 0
\end{cases}
\end{equation*}
As visualized in Figure \ref{fig:grpo_adv}, GRPO dynamically scales rewards based on prompt difficulty $\hat{p}(s_p)$. Crucially, this mechanism evaluates the \textit{entire response} as an atomic unit against a prompt-dependent baseline. Thus, although GRPO does not explicitly redefine the environment, its advantage formulation implicitly models the reasoning task as a Contextual Bandit rather than a multi-step MDP.

\vspace{0.5em}
\noindent \textbf{This observation prompts a critical question:} \textit{Does GRPO's success imply that PPO's instability stems from the token-level MDP decomposition, rather than the fundamental intractability of value estimation?}

\section{Method}
\label{subsec:sppo_method}

\subsection{Formulation: Sequence-Level Contextual Bandit}
\label{subsec:bandit_formulation}
\textbf{To answer this and formalize the implicit insight}, we explicitly reformulate the reasoning process from a token-level MDP to a \textbf{Sequence-Level Contextual Bandit (SL-CB)}. By conceptually collapsing the time horizon ($H=1$), we map the reasoning task to the tuple $(\mathcal{S}, \mathcal{A}, r)$, where the context $\mathcal{S}$ is defined strictly by the static prompt $s_p$, and the action $\mathcal{A}$ treats the \textit{entire response sequence} $a_{seq} = (y_1, \dots, y_T)$ as a single atomic unit. Accordingly, the reward $r(s_p, a_{seq})$ evaluates the \textbf{holistic correctness} of the generated chain.

This formulation fundamentally circumvents the credit assignment ambiguity inherent to MDPs. Rather than forcing a Critic to decompose sparse outcomes into noisy token-level signals, we optimize the expected sequence reward conditioned strictly on the prompt. Consequently, the value function $V(s_p)$ simplifies to estimating the scalar \textbf{solvability} of the problem, aligning directly with the objective of sparse-reward reasoning.

\subsection{SPPO: Sequence-Level Proximal Policy Optimization}
Building upon the theoretical insights of the sequence-level bandit formulation, we propose \textbf{SPPO}, an algorithm designed to strictly align the optimization objective with the sparse, outcome-oriented nature of reasoning tasks.

\paragraph{Value Function and Advantage Estimation}
First, we redefine the role of the critic. Unlike the token-level value function in standard PPO, which attempts to predict future returns from arbitrary intermediate states, we train a value model $V_\phi(s_p)$ to estimate the \textbf{scalar probability of success} for a given prompt $s_p$. 

To construct the advantage, we treat the single sample outcome as a realization of a Bernoulli trial with probability $V_\phi(s_p)$. We adopt a standardized advantage formulation to stabilize training:
\begin{equation}
A(s_p, a) = R - V_\phi(s_p)
\end{equation}
where $R \in \{0, 1\}$ is the binary reward. This formulation naturally amplifies the signal when the model is confident but wrong, and suppresses noise when the model is uncertain ($V \approx 0.5$).

To ensure $V_\phi(s_p)$ serves as a calibrated baseline, we optimize it using the Binary Cross-Entropy (BCE) loss:
\begin{equation} \label{eq:bce_loss}
\begin{split}
    L_V(\phi) = - \mathbb{E} \Big[ & R \log V_\phi(s_p) + \\
    & (1 - R) \log (1 - V_\phi(s_p)) \Big]
\end{split}
\end{equation}

\paragraph{Sequence-Level Policy Optimization}
With the advantage established, we formulate the policy optimization objective. SPPO adapts the clipped surrogate objective of PPO but fundamentally alters the scope of the advantage term. The objective function is defined as:
\begin{equation}
\label{eq:sppo_objective}
\begin{aligned}
J_{\text{SPPO}}(\theta)
&=
\mathbb{E}_{s_p \sim \mathcal{D},\, a \sim \pi_{\theta_k},\, t \in a}
\Big[
\min \big(
r_t(\theta) A(s_p, a), \\
&\qquad
\text{clip}\!\big(r_t(\theta), 1-\epsilon, 1+\epsilon\big) A(s_p, a)
\big)
\Big]
\end{aligned}
\end{equation}

Here, $r_t(\theta) = \frac{\pi_\theta(a_t | s_p, a_{<t})}{\pi_{\theta_k}(a_t | s_p, a_{<t})}$ represents the probability ratio between the current policy $\pi_\theta$ and the behavior policy $\pi_{\theta_k}$ for the token at timestep $t$, and $\epsilon$ is the standard clipping hyperparameter to constrain the policy update. 

Crucially, unlike standard PPO where each token $t$ is assigned a unique, time-dependent advantage $\hat{A}_t$ via GAE, SPPO propagates the single sequence-level advantage $A(s_p, a)$ uniformly to all constituent tokens $t$ in the sequence $a$. This mechanism ensures that if a reasoning chain leads to a correct answer ($A > 0$), every step in that chain is reinforced equally; conversely, if the chain fails ($A < 0$), every step is penalized. By decoupling the advantage signal from the sequence length, SPPO effectively solves the temporal credit assignment problem that hinders standard PPO in Long Chain-of-Thought tasks.
\section{Experiments}
\label{sec:experiments}
\subsection{Experimental Setup}
We evaluate \textbf{DeepSeek-R1-Distill-Qwen-1.5B} and \textbf{DeepSeek-R1-Distill-Qwen-7B} models, fine-tuned on DeepScaleR \citep{deepscaler2025} and DAPO-17K \citep{yu2025dapoopensourcellmreinforcement} respectively. We evaluate performance using Average@16 accuracy across five held-out benchmarks: \textbf{AIME24}\citep{aime2024}, \textbf{AIME25}\citep{aime2024}, \textbf{AMC23}\citep{amc2023}, \textbf{MATH500}\citep{hendrycksmath2021}, and \textbf{Minerva Math}\citep{lewkowycz2022solvingquantitativereasoningproblems}. Comprehensive links to all models and datasets are provided in Appendix \ref{app:resources}.
\paragraph{Baselines \& Implementation}
We benchmark SPPO against: (1) \textbf{Base Model}; (2) \textbf{Standard PPO} (token-level); and (3) Sequence-level methods including \textbf{ReMax}, \textbf{RLOO}, and \textbf{GRPO} ($N=8$). All algorithms are implemented via \texttt{verl} \citep{Sheng_2025} using outcome-based rewards ($+1$ for correct boxed answers and 0 for the incorrect answers). We utilized the precise reward function implemented in \texttt{Reasoning360} \citep{cheng2025revisitingreinforcementlearningllm} to conduct both training and evaluation.
\textbf{Hyperparameters:} We set $\beta_{KL} = 0$ to encourage exploration. Global batch sizes are configured at 256 for the 1.5B model and 512 for the 7B model. Learning rates are set to $1\text{e-}6$ for Actors and $5\text{e-}6$ for Critics. Standard PPO employs $\gamma=1, \lambda=1$ to propagate sparse rewards. The hyperparameters for all baselines generally follow the official recommended examples provided by the \texttt{verl} library. All experiments were conducted on $4 \times \text{A100}$ (1.5B) and $4 \times \text{H100}$ (7B) GPUs. For complete reproducibility, we provide the exact execution scripts and configuration commands for both SPPO and the baselines in Appendix \ref{app:implementation_details}.

\subsection{Main Results}
\label{subsec:main_results}
Table \ref{tab:main_results} presents the performance comparison. Standard PPO struggles to consistently improve over the base model, confirming the instability of GAE in sparse-reward settings. While sequence-level baselines (ReMax, RLOO) improve stability, they generally lag behind group-based approaches.

\textbf{SPPO} achieves the highest overall performance, surpassing GRPO ($N=8$) on most benchmarks (Avg 48.06 vs. 47.08 on 1.5B). Crucially, SPPO achieves this with single-sample efficiency ($N=1$), effectively eliminating the ``Tail Effect'' (Figure \ref{fig:value_charts}) without the computational bottleneck of multi-sampling.

\begin{table}[t]
    \centering
    \renewcommand{\arraystretch}{1.1} 
    \setlength{\tabcolsep}{3.5pt}
    \resizebox{\columnwidth}{!}{
        \begin{tabular}{lcccccc}
        \toprule
        \textbf{Method} & \textbf{A24} & \textbf{A25} & \textbf{AMC} & \textbf{MATH} & \textbf{Minerva} & \textbf{Avg} \\
        \midrule
        \multicolumn{7}{c}{\textit{\textbf{DeepSeek-R1-Distill-Qwen-1.5B}}} \\
        \midrule
        Base Model   & 27.50 & 21.67 & 71.56 & 83.73 & 20.35 & 44.96 \\
        PPO          & 27.50 & 20.83 & 70.63 & 81.38 & 19.89 & 44.06 \\
        ReMax        & \underline{31.67} & 25.42 & 71.88 & \textbf{84.38} & 20.40 & 46.74 \\
        RLOO         & 30.42 & 21.67 & 72.81 & \underline{84.10} & 21.73 & 46.15 \\
        GRPO ($N=8$) & 30.00 & \textbf{26.25} & \underline{73.13} & 83.88 & \textbf{22.15} & \underline{47.08} \\
        \rowcolor{gray!10} \textbf{SPPO (Ours)} & \textbf{34.17} & \underline{25.83} & \textbf{74.38} & 83.78 & \textbf{22.15} & \textbf{48.06} \\
        
        \midrule
        \multicolumn{7}{c}{\textit{\textbf{DeepSeek-R1-Distill-Qwen-7B}}} \\
        \midrule
        Base Model   & 41.25 & 26.67 & 79.38 & 87.20 & 27.94 & 52.49 \\
        PPO          & 45.20 & \textbf{35.42} & 85.31 & 88.48 & 27.80 & 56.44 \\
        ReMax        & 49.38 & 31.25 & 86.56 & \underline{90.28} & 27.99 & 57.09 \\
        RLOO         & 46.67 & 32.50 & \underline{86.88} & \textbf{90.35} & 28.72 & 57.02 \\
        GRPO ($N=8$) & 47.08 & \underline{35.00} & 86.25 & 90.15 & \underline{28.74} & 57.44 \\
        \rowcolor{gray!10} \textbf{SPPO (Ours)} & \underline{50.83} & \underline{35.00} & 86.25 & 90.13 & 28.35 & \underline{58.11} \\
        \rowcolor{gray!10} \hspace{0.5em}\textit{w/ Small Critic} & \textbf{52.29} & 34.58 & \textbf{87.19} & 89.88 & \textbf{28.86} & \textbf{58.56} \\
        \bottomrule
        \end{tabular}
    }
    \caption{Performance comparison on 1.5B and 7B scales. SPPO consistently outperforms baselines. The \textit{Small Critic} variant (1.5B Critic aligning 7B Policy) achieves the top average score.}
    \label{tab:main_results}
\end{table}

\paragraph{Critic Decoupling}
We hypothesize that scalar solvability estimation is significantly simpler than generative reasoning, permitting a smaller Value Function. To test this, we trained the 7B Policy using a \textbf{1.5B Critic}. As shown in Table \ref{tab:main_results} (\textit{w/ Small Critic}), this configuration not only retains effectiveness but achieves the highest average score (58.56). This validates that a lightweight critic can effectively align a large policy, significantly reducing the memory footprint of RLVR training (Figure \ref{fig:memory_usage}).

\subsection{Ablation Study: Impact of Loss Function}
\label{subsec:ablation_loss}
To isolate the source of SPPO's performance gains, we ablated the architectural contribution by applying the BCE loss to the standard token-level PPO framework (\textbf{PPO + BCE}). 

\begin{figure}[h]
    \centering
    \includegraphics[width=1.0\columnwidth]{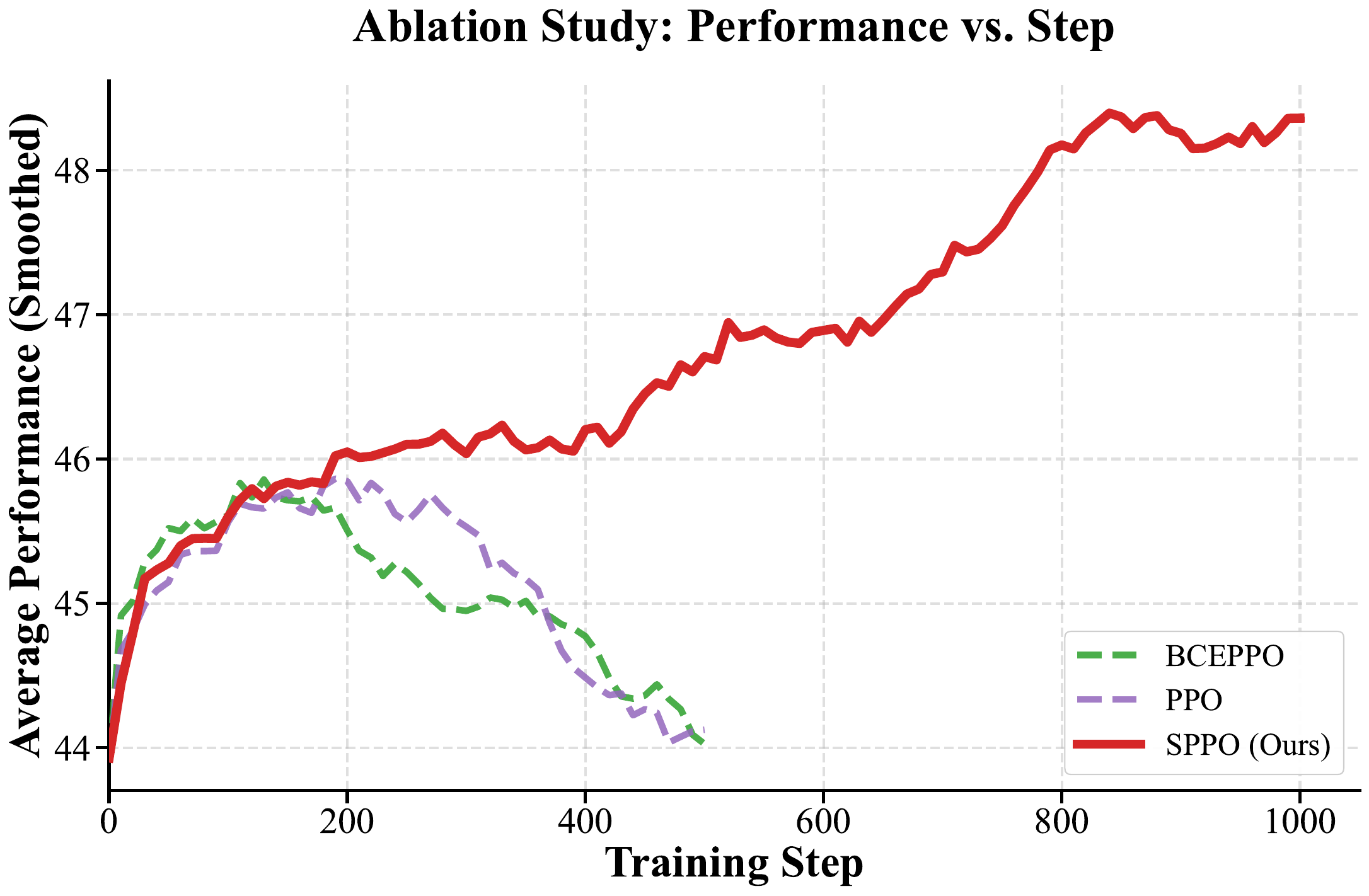}
    \caption{\textbf{Ablation Analysis of the Optimization Objective.} We compare SPPO against Standard PPO and a control baseline (``PPO + BCE'') that integrates the BCE loss into the token-level framework. The failure of the control baseline demonstrates that the performance gains do not stem from the loss function itself, but from the \textbf{Sequence-Level Contextual Bandit formulation}, which propagates a unified advantage signal to resolve credit assignment ambiguity.}
    \label{fig:ablation_loss}
\end{figure}

As shown in Figure \ref{fig:ablation_loss}, \textbf{PPO + BCE} fails to reproduce the success of SPPO, exhibiting the same instability as the standard baseline. Notably, we terminated both PPO-based runs early at 500 steps due to observed performance collapse and degrading scores. This empirical evidence validates that SPPO's efficacy derives fundamentally from its \textbf{Sequence-Level Contextual Bandit formulation}—specifically the propagation of a unified advantage signal $A = R - V(s_p)$—rather than the adoption of the BCE loss in isolation.

\section{Analysis}
\subsection{Scalability and Computational Efficiency}
\label{subsec:scalability}

\begin{figure}[t]
    \centering
    \includegraphics[width=1.0\columnwidth]{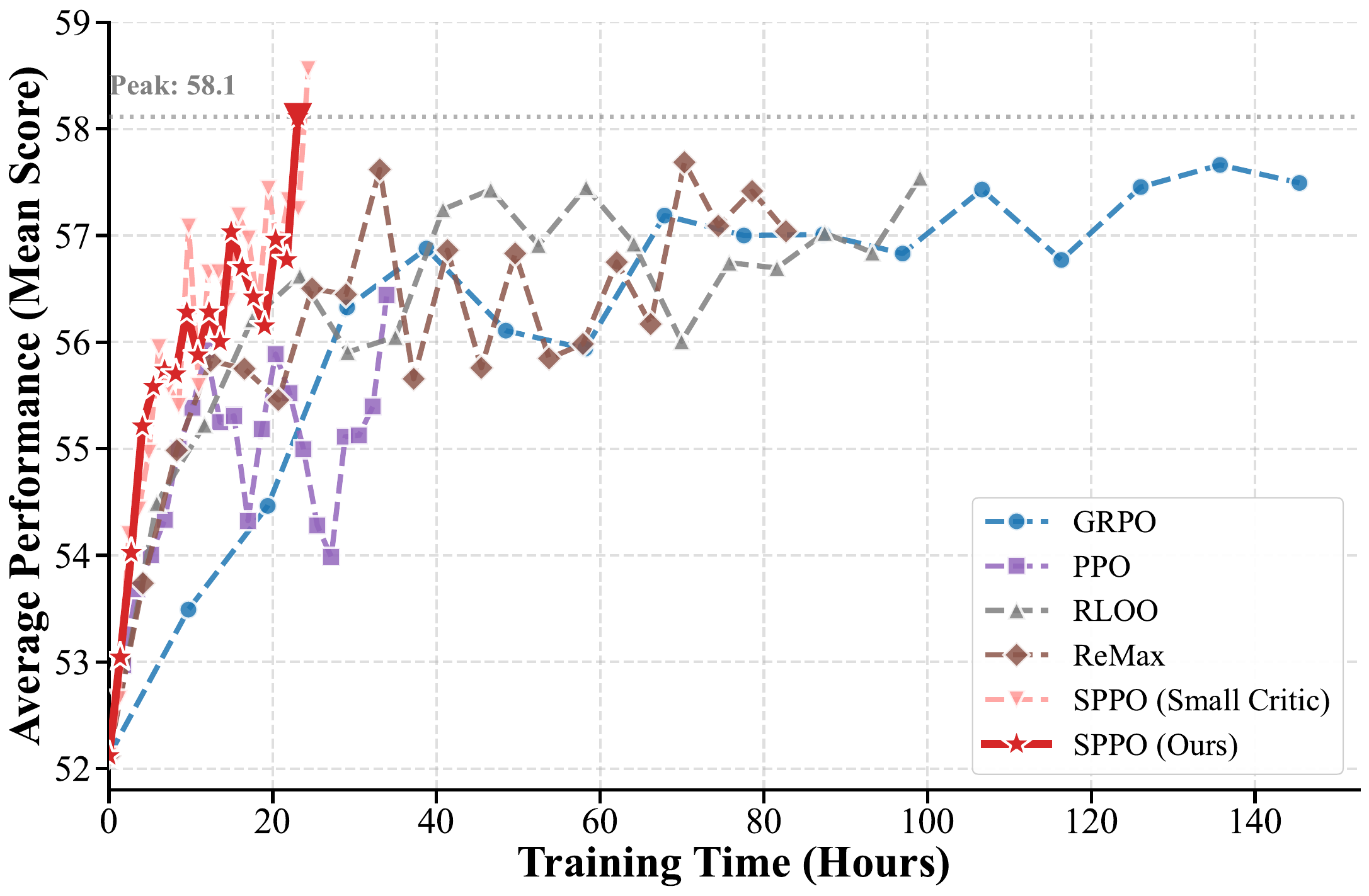}
    \caption{\textbf{Training Efficiency on Deepseek-R1-Distill-Qwen-7B (Performance vs. Wall-clock Time).} The plot compares the trajectory of SPPO against strong baselines (GRPO, PPO, RLOO, ReMax) on the DAPO-17k dataset\citep{yu2025dapoopensourcellmreinforcement}. \textbf{Solid Red:} SPPO with a matched 7B Critic. \textbf{Dashed Pink:} SPPO with a decoupled, smaller 1.5B Critic (Deepseek-R1-Distill-Qwen-1.5B). The y-axis noted as the Avg@8 score evaluated on \textbf{AIME24}, \textbf{AIME25}, \textbf{AMC23}, \textbf{MATH500}, and \textbf{Minerva Math}. SPPO achieves optimal performance significantly faster than group-based methods, and the decoupled critic maintains performance while reducing memory overhead.}
    \label{fig:efficiency_plot}
\end{figure}

To evaluate the scalability of SPPO and its efficiency in larger parameter regimes, we extended our experiments to the \textbf{DeepSeek-R1-Distill-Qwen-7B} model. For this analysis, we utilized the DAPO-17K dataset \cite{deepscaler2025}, a curated subset of high-quality mathematical reasoning problems. We compared SPPO against the baseline algorithms, tracking performance on the hold-out validation set against wall-clock training time.

\paragraph{Training Efficiency}
As illustrated in Figure \ref{fig:efficiency_plot}, SPPO demonstrates superior training efficiency compared to group-based alternatives. GRPO ($N=8$) and RLOO exhibit a slower ``time-to-convergence'' primarily due to the computational bottleneck of generating multiple samples per prompt to estimate the baseline. In contrast, SPPO, which operates with single-sample efficiency ($N=1$), updates the policy more frequently within the same time window. Consequently, SPPO reaches peak performance (mean score $\approx 58$) in approximately 22 hours, whereas baselines require significantly longer to reach comparable levels or plateau at lower scores (e.g., standard PPO).

\begin{figure}[t]
    \centering
    \includegraphics[width=1.0\columnwidth]{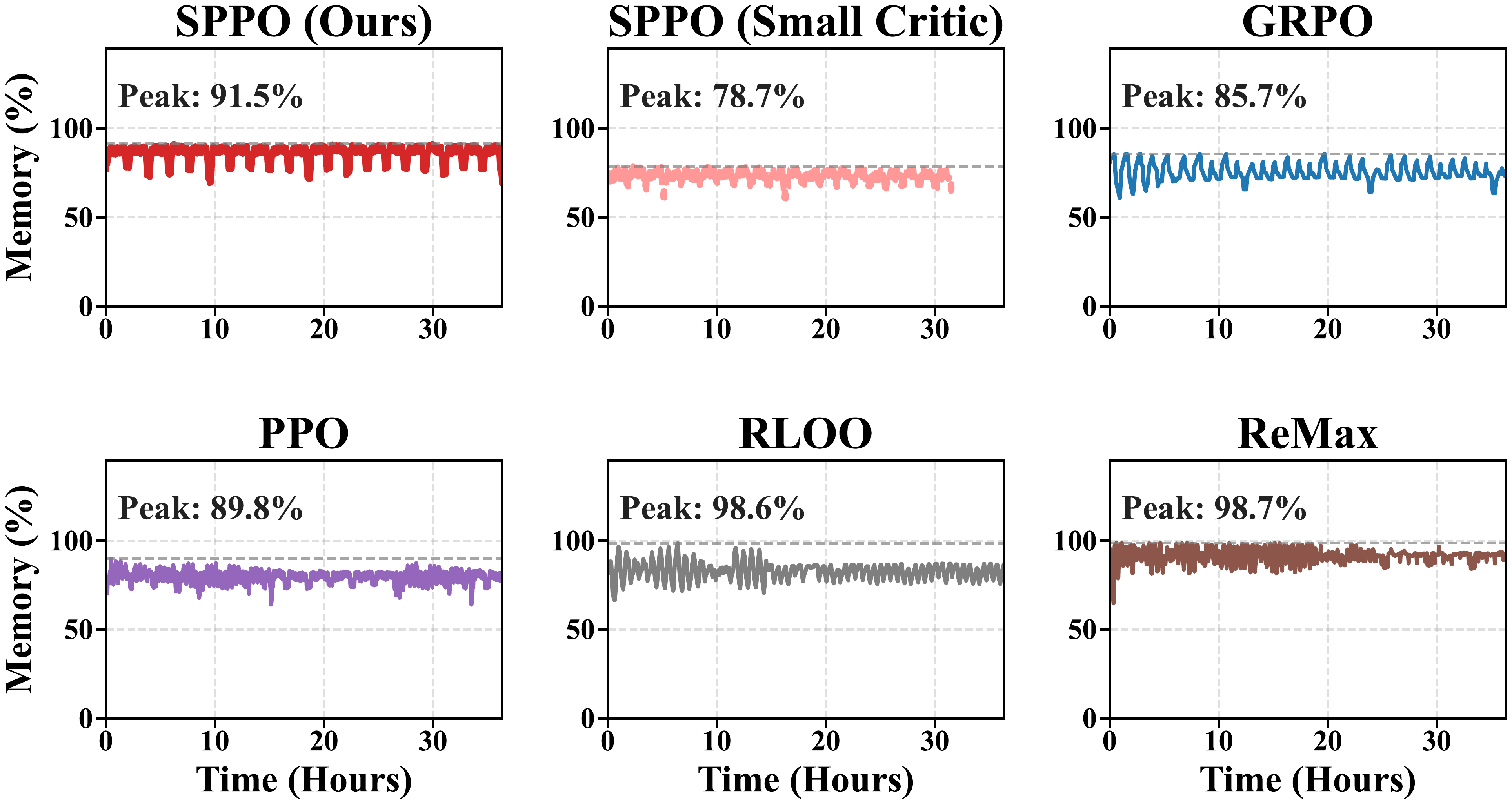}
    \caption{\textbf{GPU Memory Allocation Analysis.} Comparison of normalized peak VRAM usage during the training of a 7B policy. The ``Decoupled Critic'' (7B+1.5B) approach, combined with the system-level optimizations in \texttt{verl}, significantly reduces memory bottlenecks compared to symmetric actor-critic setups (7B+7B), making efficient RL alignment accessible on consumer-grade hardware.}
    \label{fig:memory_usage}
\end{figure}

\paragraph{Resource Efficiency and VRAM Optimization}
Beyond computational throughput, the ``Small Critic'' configuration offers a decisive advantage in hardware accessibility. Standard RLHF typically requires loading a Critic of equal size to the Policy, effectively doubling the parameter memory footprint. By decoupling the critic size (1.5B) from the policy (7B), SPPO significantly alleviates this bottleneck. Furthermore, by leveraging the advanced memory management and sharding techniques provided by the \texttt{verl} library \citep{Sheng_2025}, our implementation minimizes redundant memory allocation. 

As visualized in Figure \ref{fig:memory_usage}, the SPPO framework with a decoupled critic maintains a low memory profile. This confirms that SPPO is not only algorithmically stable but also resource-efficient, enabling the alignment of large reasoning models even under constrained GPU budgets.

\subsection{Value Model Analysis: Calibration and Correlation}
\label{subsec:critic_analysis}

The efficacy of SPPO relies heavily on the quality of the sequence-level value function, $V(s_p)$. In our framework, $V(s_p)$ serves as the baseline for advantage estimation ($A = R - V_\phi(s_p)$). Theoretically, an ideal value model should accurately capture the intrinsic difficulty of a prompt, approximating the expected success rate of the current policy. To validate this, we conducted a correlation analysis between the critic's predictions and the empirical ground truth on a held-out validation set.

\paragraph{Setup}
We randomly sampled a diverse set of $N=200$ prompts and executed the policy multiple times for each to compute the empirical pass rate (AVG@$k$), which serves as the ground truth label for difficulty. We then compared these empirical values against the predicted probabilities output by our Value Model.

\begin{figure}[htbp]
  \centering
  \includegraphics[width=1.0\columnwidth]{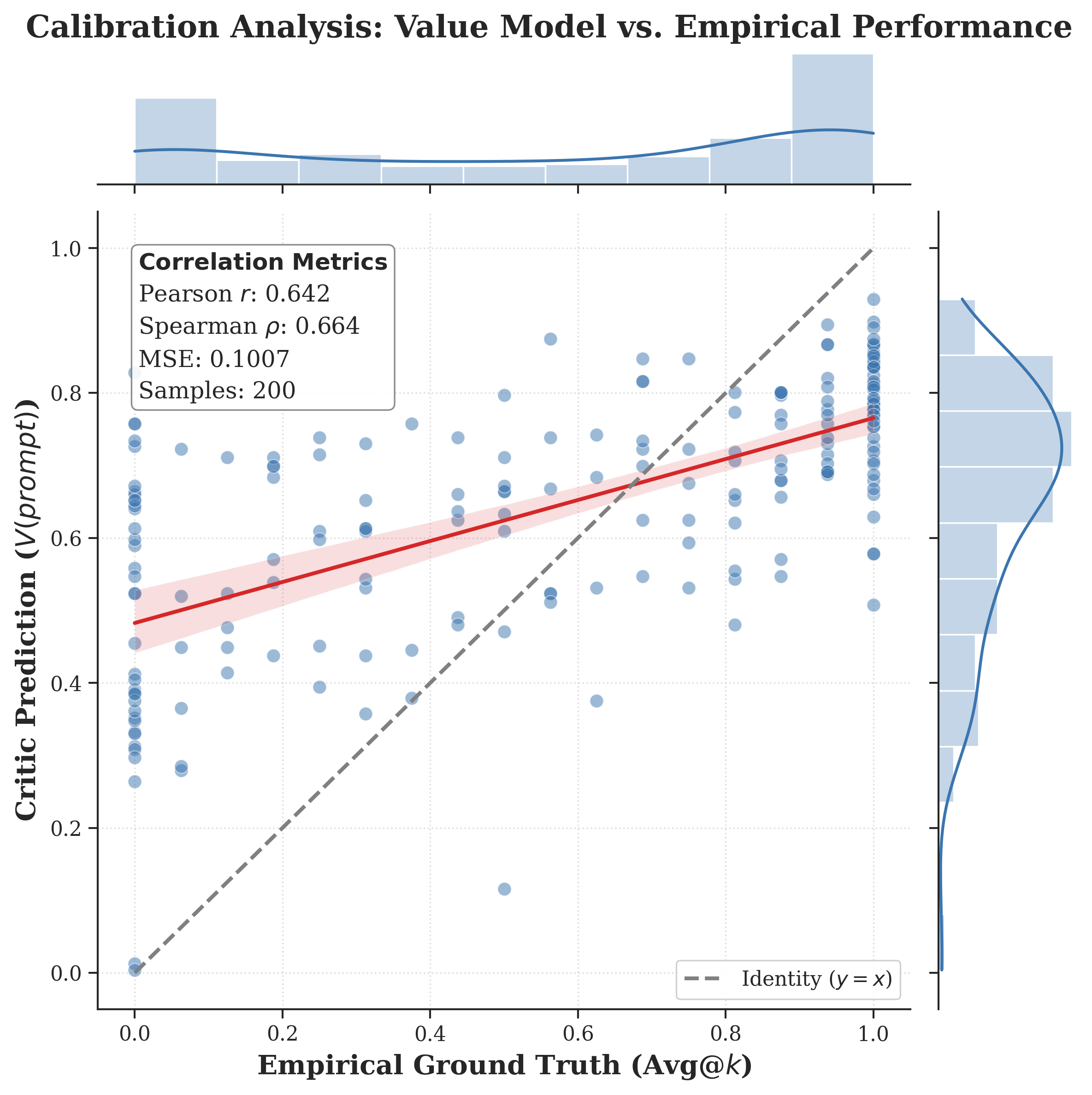}
  \caption{Correlation analysis between the Critic's predicted difficulty ($y$-axis) and the empirical AVG@$k$ rate ($x$-axis) (Where k = 64). The plot reveals a clear positive correlation (Pearson $r=0.642$), indicating the Critic successfully distinguishes between hard and easy tasks. The marginal histograms (top and right) contrast the bimodal distribution of real task difficulty with the more conservative, quasi-normal distribution of the Critic's predictions.}
  \label{fig:critic_correlation}
\end{figure}

\paragraph{Calibration and Distribution Analysis}
As illustrated in Figure \ref{fig:critic_correlation}, our analysis reveals a distinct positive linear correlation between the predicted values and the empirical results. With a Pearson correlation coefficient of $\mathbf{0.642}$ and a Spearman rank correlation of $\mathbf{0.664}$, we find strong statistical evidence that the Value Model has successfully learned to capture the relative difficulty of prompts, effectively cutting through the inherent stochasticity of LLM generation. Beyond these correlation metrics, the marginal histograms provide a critical insight into the model's behavioral tendencies: while the empirical difficulty (top histogram) follows a \textbf{bimodal distribution}—where tasks are typically either completely unsolvable (0.0) or consistently solvable (1.0)—the critic's predictions (right histogram) exhibit a \textbf{unimodal, quasi-normal distribution} centered around 0.6-0.7. This distributional shift indicates that the critic adopts a conservative prediction strategy, aggregating uncertainty rather than overfitting to the binary extremes of the empirical data.

\paragraph{Implications for SPPO}
The distributional discrepancy suggests that the Critic tends to be \textit{conservative}, exhibiting a ``regression to the mean'' behavior rather than predicting extreme probabilities (0 or 1). However, the regression trend (red line) maintains a clear positive slope. This confirms that $V(s_p)$ serves as a valid, variance-reducing baseline. 

Specifically, for hard prompts (Avg@$k \approx 0$), the Critic predicts lower values ($\approx 0.5$), ensuring that a rare success yields a strong positive advantage. For easy prompts (Avg@$k \approx 1$), the Critic predicts higher values ($\approx 0.8$), ensuring that a failure yields a significant negative penalty.
\subsection{Controlled Analysis: The RLVR Benchmark}
\label{subsec:benchmark_experiment}

To strictly disentangle the algorithmic efficacy of SPPO from system-level optimizations inherent to the \texttt{verl} framework, and to rigorously validate its robustness in isolation, we extend our evaluation to a suite of \textbf{five} representative control environments: \textit{Precision CartPole}, \textit{MountainCar}, \textit{Hopper} (MuJoCo), \textit{LunarLander}, and \textit{Pendulum}. We reconfigure these classic control tasks into a \textbf{Reinforcement Learning with Verifiable Rewards (RLVR)} framework. By strictly enforcing structural constraints—specifically long time horizons, deterministic transitions, and sparse binary outcome feedback—we construct a minimalist testbed that mimics the optimization landscape of LLM reasoning without the confounding variables of large-scale distributed training.

\paragraph{Experimental Protocol}
To mirror the LLM alignment lifecycle, we implement a rigorous three-stage pipeline. First, \textbf{Expert Synthesis} trains policies using dense, shaped rewards (e.g., velocity bonuses in \textit{MountainCar}, upright incentives in \textit{Pendulum}) to mimic pre-training supervision. Subsequently, \textbf{Supervised Fine-Tuning (SFT)} applies behavior cloning to filtered successful trajectories (e.g., $r>0.5$ or state $x > 1.0$ in \textit{Hopper}), initializing a model with non-zero but imperfect solvability. Finally, \textbf{RL Fine-tuning} introduces the core sparse reward challenge: agents receive a strictly binary terminal reward $r_H \in \{0, 1\}$ with zero intermediate feedback ($r_t = 0$ for $t < H$) and a discount factor $\gamma = 1.0$, compelling the algorithm to bridge the full temporal horizon.

\paragraph{Task Configurations and Hyperparameters}
To ensure a fair comparison of optimization objectives, we align core PPO hyperparameters (e.g., $\epsilon=0.2$, learning rates) across algorithms while adapting batch sizes to the distinct exploration dynamics of each domain. Specifically, we utilize a strictly aligned batch size of 16 trajectories per update for \textit{LunarLander}  and \textit{Pendulum}; a reduced batch size of 8 for exploration-heavy tasks (\textit{MountainCar}, \textit{Hopper}); and a batch size of 64 for the rapid-dynamics \textit{CartPole}. Success in the RL phase is determined by rigorous outcome criteria: \textit{Precision CartPole} ($H=200$) requires a final angle $|\theta| \le 0.5^\circ$; \textit{MountainCar} ($H=1000$) requires reaching the flag ($x \ge 0.45$); \textit{Hopper} ($H=1000$) demands survival with forward progress ($x > 1.0$ m); \textit{LunarLander} ($H=1000$) necessitates stable leg contact ($>0.5$) within the landing pad ($|x| < 0.4$); and \textit{Pendulum} ($H=1000$) requires an upright final position ($\cos \theta > 0.8$).

\begin{figure}[t]
  \centering
  \includegraphics[width=1.0\columnwidth]{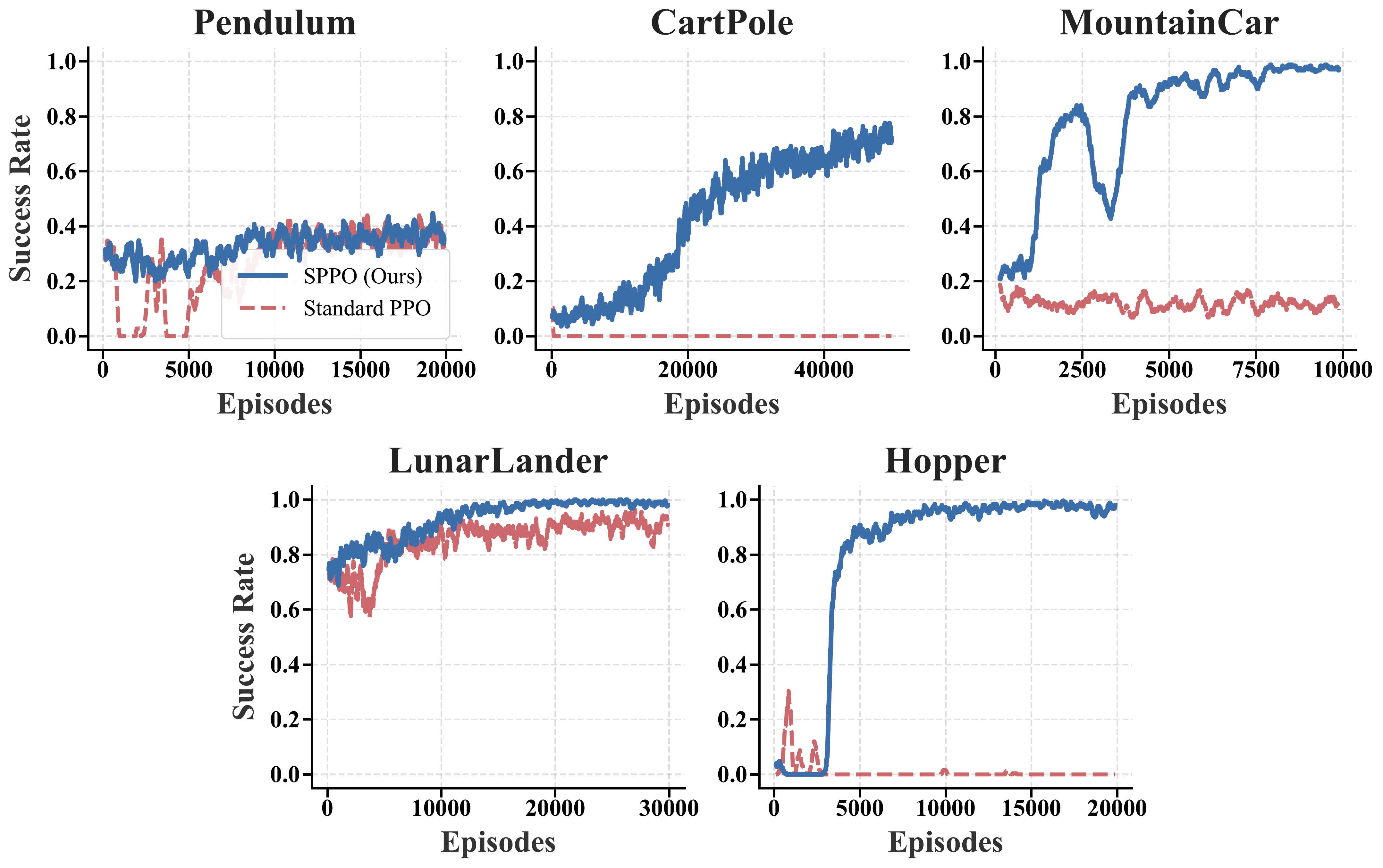}
  \caption{\textbf{RLVR Benchmark Results.} Comparison of SPPO (Blue Solid) and Standard PPO (Red Dashed) across five control tasks with sparse outcome rewards ($\gamma=1.0$). SPPO demonstrates robust convergence in complex control tasks where Standard PPO exhibits instability or failure (e.g., \textit{Hopper}, \textit{MountainCar}), while achieving superior sample efficiency in precision tasks like \textit{CartPole}.}
  \label{fig:benchmark_5}
\end{figure}

\paragraph{Results and Analysis}
As illustrated in Figure \ref{fig:benchmark_5}, SPPO consistently matches or outperforms Standard PPO across all domains, confirming its structural superiority in sparse-reward settings. In long-horizon tasks like \textit{Hopper} ($H=1000$) and \textit{MountainCar}, where the SFT initialization provides a weak prior, Standard PPO flatlines near zero (or stays at a low success rate) as the token-level critic $V(s_t)$ fails to propagate the sparse signal effectively; conversely, SPPO successfully solves these tasks by estimating the sequence-level solvability $V(s_0)$. Furthermore, in \textit{LunarLander}, SPPO maintains monotonic improvement, avoiding the instability observed in the Standard PPO baseline. Finally, SPPO demonstrates superior precision alignment in \textit{Precision CartPole}, rapidly converging to high-precision behaviors where step-level attribution struggles to differentiate between ``good'' and ``perfect'' trajectories given binary feedback.
\section{Related Work}
\label{related work}
\subsection{Reinforcement Learning Algorithms in LLMs}
\label{rel:RL4LLM}

Proximal Policy Optimization (PPO) \citep{schulman2017proximalpolicyoptimizationalgorithms} aligns LLMs using a dense, token-level value function. However, in sparse-reward reasoning tasks (RLVR), this approach struggles as GAE fails to effectively assign credit over long Chain-of-Thought horizons \citep{lightman2023letsverifystepstep}.
Group-based methods like GRPO \citep{shao2024deepseekmathpushinglimitsmathematical} mitigate this by estimating baselines via multi-sampling ($N>1$). By normalizing rewards against the group mean, GRPO implicitly adopts a sequence-level objective, bypassing temporal credit assignment issues.

While recent variants like DAPO \citep{deepscaler2025} and Dr.GRPO \citep{liu2025understandingr1zeroliketrainingcritical} propose strategies to refine gradient dynamics (e.g., dynamic sampling), they remain fundamentally bound to the computationally expensive multi-sampling paradigm. \textbf{In this work, we exclude such orthogonal optimizations.} Our primary objective is to isolate and validate the effectiveness of the \textbf{Sequence-Level Contextual Bandit formulation} itself, rather than remedying the inherent instabilities of group-relative baselines.
Consequently, SPPO replaces the empirical baseline with a learned scalar value function $V(s_p)$. This enables stable, on-policy learning with single-sample efficiency ($N=1$), harmonizing PPO's throughput with the structural stability of sequence-level modeling.

\subsection{Sequence-Level Exploration}
\label{rel:SLE}
Prior research has extensively explored sequence-level Reinforcement Learning (RL) algorithms. The RLOO (REINFORCE Leave-One-Out) algorithm \citep{ahmadian2024basicsrevisitingreinforcestyle} posits that token-level modeling is often superfluous, criticizing the token-level optimization inherent in PPO. However, RLOO is established upon the REINFORCE \citep{NIPS1999_464d828b} algorithm. Recent work \citep{wang2025aspoasymmetricimportancesampling} has shown that the clipping term plays a crucial role in learning stability through a Token Masking mechanism. Furthermore, RLOO's rolling mechanism increases computational requirements as CoT trajectories lengthen. 

Moreover, GSPO (Group Sequence Policy Optimization, \citet{zheng2025groupsequencepolicyoptimization}) and GMPO (Geometric-Mean Policy Optimization, \citet{zhao2025geometricmeanpolicyoptimization}) have argued that a sequence-level reward is incongruent with PPO's token-level design. However, since these methods explicitly position their core contribution as addressing the routing instability inherent to Mixture-of-Experts (MoE) architectures, we exclude them from our baselines to maintain a focus on general reasoning alignment.

\section{Conclusion}
\label{sec:conclusion}
To resolve the trade-off between standard PPO's high-bias credit assignment and GRPO's high-variance inefficiency, we introduce SPPO, which reformulates reasoning as a Sequence-Level Contextual Bandit. By employing a scalar critic for advantage estimation, SPPO secures optimization stability with high-throughput single-sample efficiency, offering a scalable paradigm for sparse-reward tasks.
\clearpage
\newpage

\section*{Limitations}
In this work, we primarily focus on RLVR, showing that SPPO effectively harmonizes sample efficiency with structural stability in sparse-reward settings. However, our approach is explicitly tailored for tasks with verifiable outcomes to estimate prompt solvability. As a result, extending this sequence-level bandit formulation to open-ended generation tasks, which lack objective ground-truth verifiers, remains a direction for future research.

\section*{Ethical Considerations}
Our study is conducted in controlled, text-only benchmark environments and does not involve human subjects or the collection of personal data. As with other agentic and world-modeling capabilities, misuse (e.g., enabling harmful or deceptive behavior) and bias propagation are possible; we encourage responsible deployment with appropriate safeguards and oversight.

\bibliography{custom}

\appendix

\section{Derivation and Analysis of GRPO Advantage}
\label{app:grpo_derivation}

We model the $N$ sampled responses for a prompt $s_p$ as independent Bernoulli trials with success probability $p = P(R=1 | s_p)$. The GRPO advantage is defined as $Adv = (R_i - \mu_g) / \sigma_g$.

For binary rewards $R_i \in \{0, 1\}$, the sample mean corresponds to the empirical success rate $\mu_g = \hat{p}$. Using standard results for Bernoulli distributions, the unbiased sample standard deviation $\sigma_g$ converges to the population standard deviation for large $N$:
\begin{equation}
    \sigma_g = \sqrt{\frac{N}{N-1} \hat{p}(1-\hat{p})} \approx \sqrt{\hat{p}(1-\hat{p})}
\end{equation}
Substituting these into the advantage formulation, we obtain the standardized residual:
\begin{equation}
    Adv(s_p, R) \approx \frac{R - \hat{p}}{\sqrt{\hat{p}(1-\hat{p})}}
\end{equation}
Evaluation for the two possible outcomes $R \in \{0, 1\}$ yields:
\begin{equation}
Adv(s_p, a) = 
\begin{cases}
\sqrt{\frac{1-\hat{p}}{\hat{p}}} & \text{if } R = 1 \text{ (Success)} \\
-\sqrt{\frac{\hat{p}}{1-\hat{p}}} & \text{if } R = 0 \text{ (Failure)}
\end{cases}
\end{equation}

\section{Extended Visualization of Critic Dynamics}
\label{app:value_vis}

\begin{figure}[h]
    \centering
    \includegraphics[width=1.0\columnwidth]{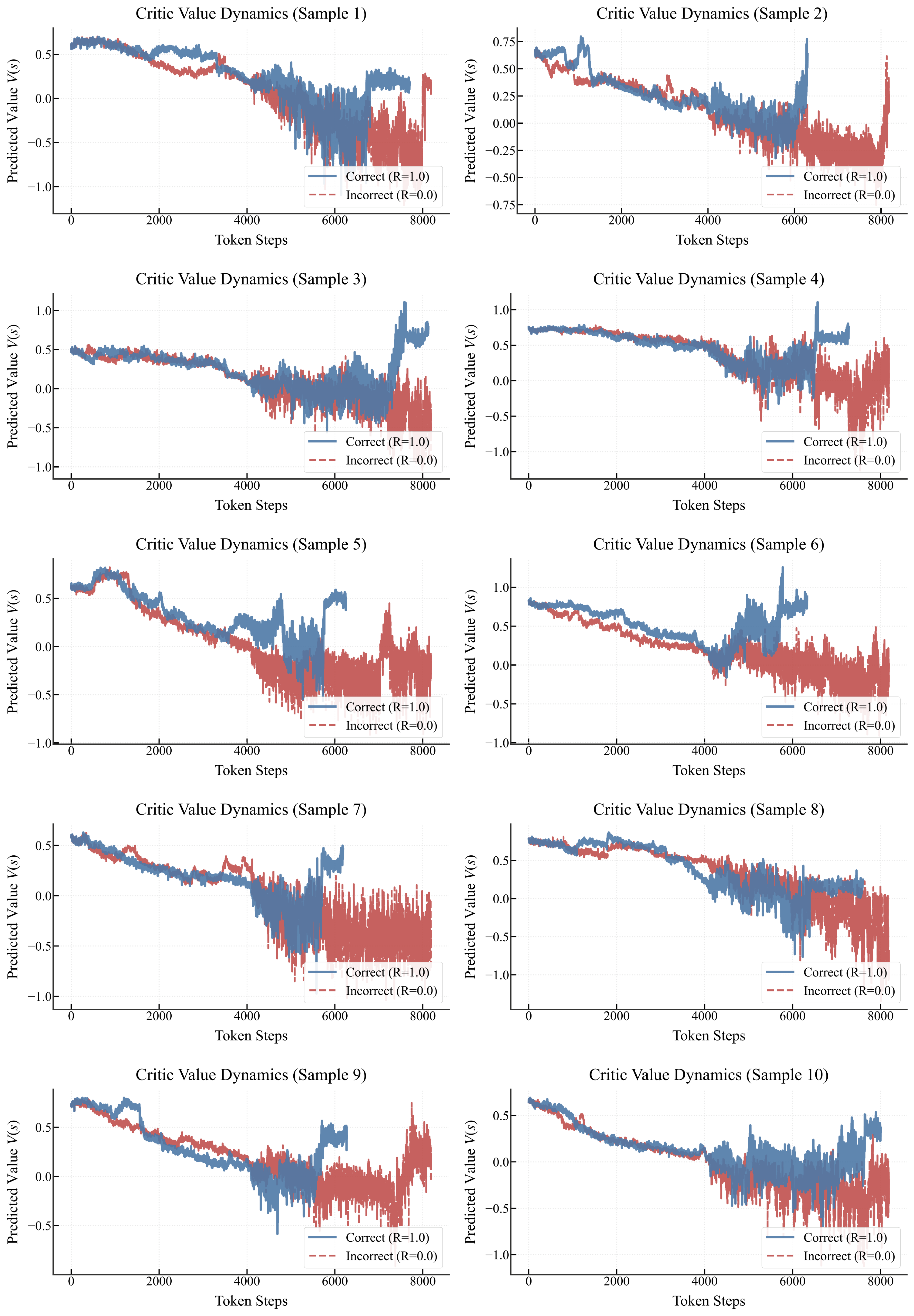}
    \caption{\textbf{Extended Analysis of Critic Value Dynamics (10 Random Samples).} 
    Each subplot represents a distinct mathematical problem sampled from the validation set. 
    \textbf{Blue Lines:} Value estimates for correct trajectories ($R=1$). 
    \textbf{Red Lines:} Value estimates for incorrect trajectories ($R=0$). 
    The consistent overlap of value curves until the sequence tail demonstrates the systematic failure of token-level value estimation in distinguishing intermediate reasoning quality.}
    \label{fig:appendix_all}
\end{figure}

In this section, we present a comprehensive visualization containing ten randomly sampled problems from the DeepScaleR dataset. Figure \ref{fig:appendix_all} plots the Critic's value estimates $V(s_t)$ over time for both correct and incorrect responses.

These samples consistently exhibit the ``Tail Effect'' analyzed in Section \ref{intro}. Across diverse reasoning tasks, the token-level Critic fails to distinguish between correct (Blue) and incorrect (Red) trajectories during the intermediate reasoning process. The value curves typically remain entangled until the final few tokens, confirming that the standard PPO Critic struggles to assign precise temporal credit in long-horizon reasoning tasks.

\section{Risks}

Our proposed SPPO algorithm relies on the assumption of verifiable outcome rewards to estimate prompt solvability. A potential risk involves the overgeneralization of this method to tasks lacking objective ground truths, such as ethical decision-making or subjective content generation. Applying sequence-level optimization in these areas without robust reward modeling may amplify biases present in the base model or lead to the generation of plausible but factually incorrect reasoning chains (hallucination). Furthermore, as SPPO lowers the computational barrier for training strong reasoning models, there is a need for continued monitoring to ensure these accessible capabilities are not deployed for generating harmful content or automating malicious tasks.

\section{Resources and Implementation Details}
\label{app:resources}

To facilitate reproducibility, we summarize the models, datasets, and benchmarks used in our experiments in Table \ref{tab:resources}. All resources are accessible via HuggingFace.

\begin{table}[h]
    \centering
    \scriptsize
    \renewcommand{\arraystretch}{1.2}
    \resizebox{\columnwidth}{!}{
    \begin{tabular}{@{}lll@{}}
    \toprule
    \textbf{Category} & \textbf{Resource / Link} & \textbf{License} \\
    \midrule
    \textbf{Models} & \href{https://huggingface.co/deepseek-ai/DeepSeek-R1-Distill-Qwen-1.5B}{DeepSeek-R1-Distill-Qwen-1.5B} & MIT \\
     & \href{https://huggingface.co/deepseek-ai/DeepSeek-R1-Distill-Qwen-7B}{DeepSeek-R1-Distill-Qwen-7B} & MIT \\
    \midrule
    \textbf{Training} & \href{https://huggingface.co/datasets/BytedTsinghua-SIA/DAPO-Math-17k}{DAPO-Math-17k} & Apache 2.0 \\
    \textbf{Datasets} & \href{https://huggingface.co/datasets/agentica-org/DeepScaleR-Preview-Dataset}{DeepScaleR-Preview} & MIT \\
    \midrule
    \textbf{Benchmarks} & \href{https://huggingface.co/datasets/HuggingFaceH4/aime_2024}{AIME 2024}, \href{https://huggingface.co/datasets/yentinglin/aime_2025}{AIME 2025} & -- \\
     & \href{https://huggingface.co/datasets/math-ai/amc23}{AMC 23}, \href{https://huggingface.co/datasets/HuggingFaceH4/MATH-500}{MATH-500} & -- \\
     & \href{https://huggingface.co/datasets/math-ai/minervamath}{Minerva Math} & -- \\
    \bottomrule
    \end{tabular}
    }
    \caption{Summary of resources and licenses used in this work. Click on the resource names to visit their HuggingFace pages.}
    \label{tab:resources}
\end{table}
\clearpage
\onecolumn
\section{Implementation Details and Execution Commands}
\label{app:implementation_details}

In this section, we provide snapshots of the exact execution commands used to reproduce the experimental results. The images are rendered from the actual training scripts.

\subsection{SPPO (Ours)}

{
    \centering
    \includegraphics[width=0.85\textwidth]{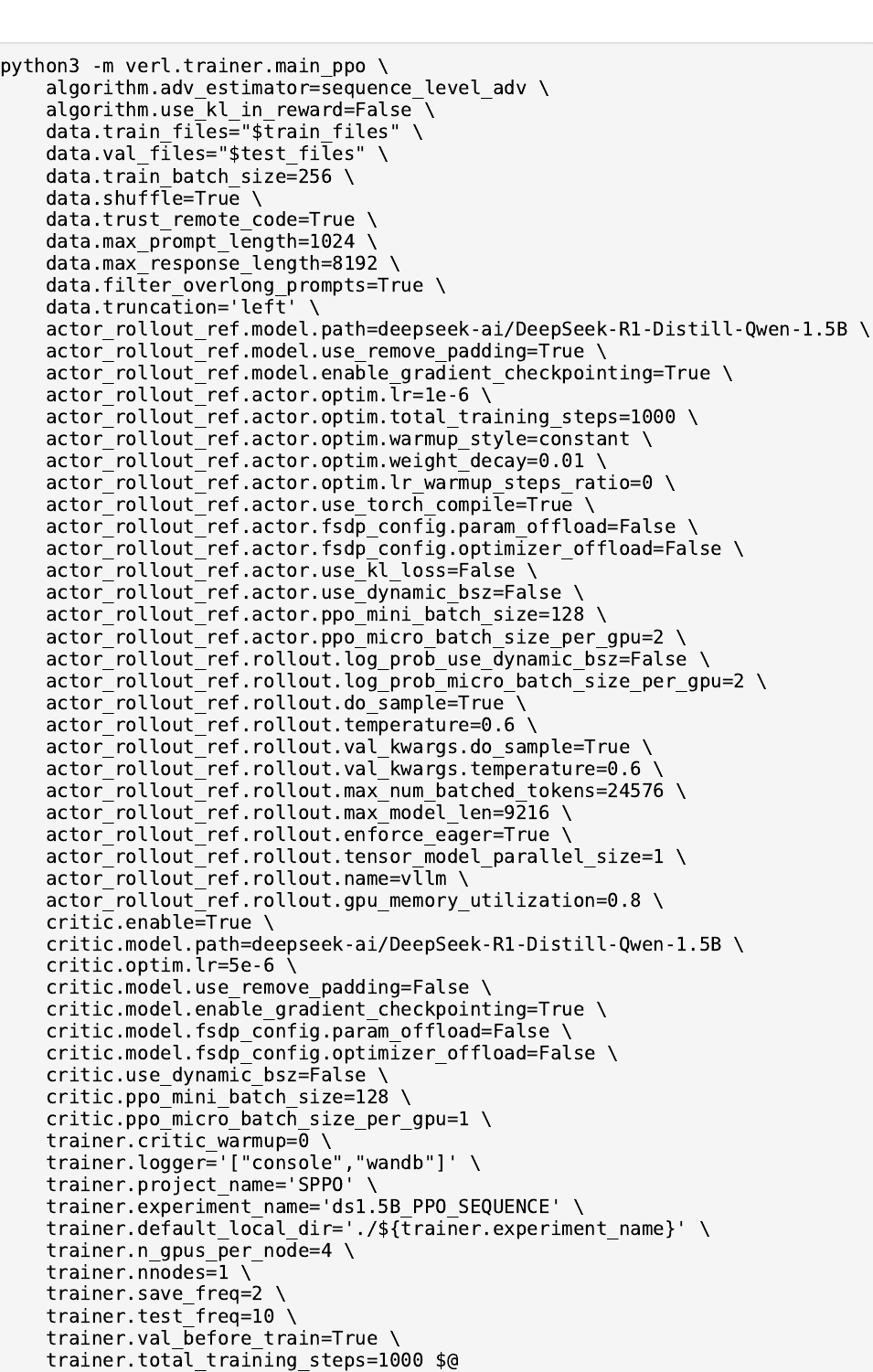}
    \captionof{figure}{Execution Command: SPPO 1.5B (Symmetric)}
    \label{fig:cmd_sppo_1.5b}
}
\vspace{1em}

{
    \centering
    \includegraphics[width=0.90\textwidth]{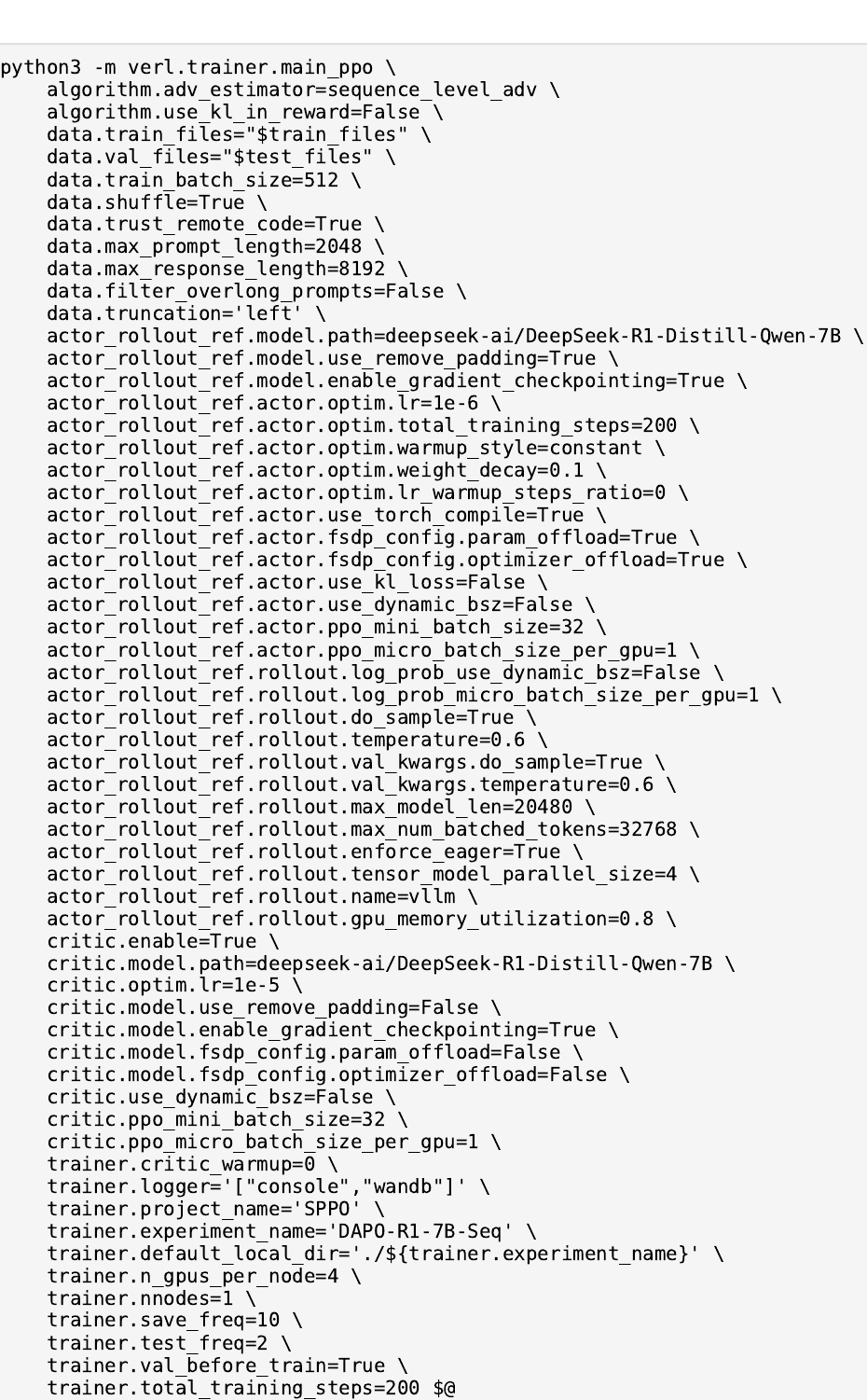}
    \captionof{figure}{Execution Command: SPPO 7B (Symmetric)}
    \label{fig:cmd_sppo_7b}
}
\vspace{1em}

{
    \centering
    \includegraphics[width=0.90\textwidth]{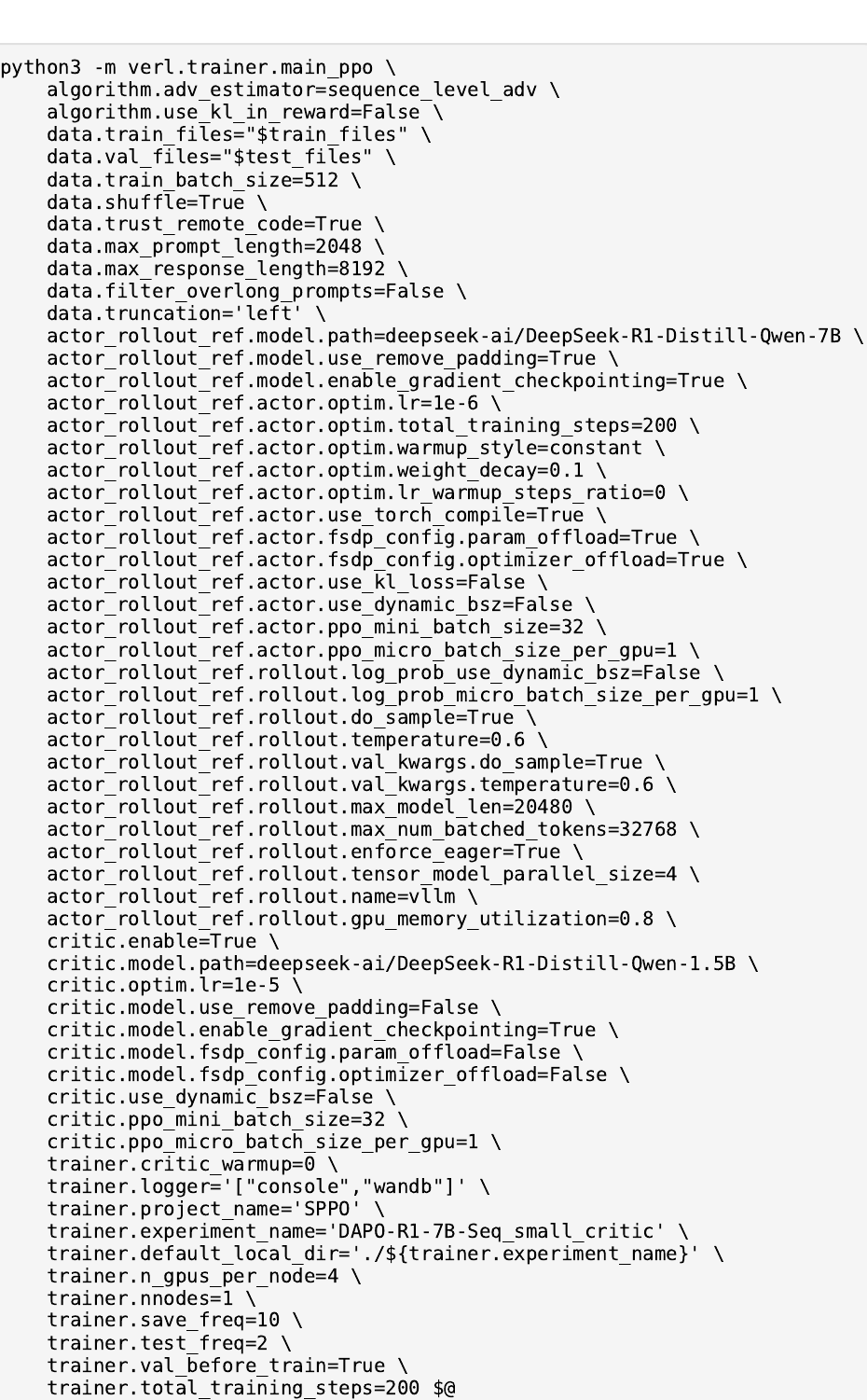}
    \captionof{figure}{Execution Command: SPPO 7B (Decoupled / Small Critic)}
    \label{fig:cmd_sppo_7b_small}
}
\vspace{2em}

\noindent
\begin{minipage}{\textwidth}
    \subsection{Group Relative Policy Optimization (GRPO)}
    
    \vspace{0.5em}

    {
        \centering
        \includegraphics[width=0.95\textwidth]{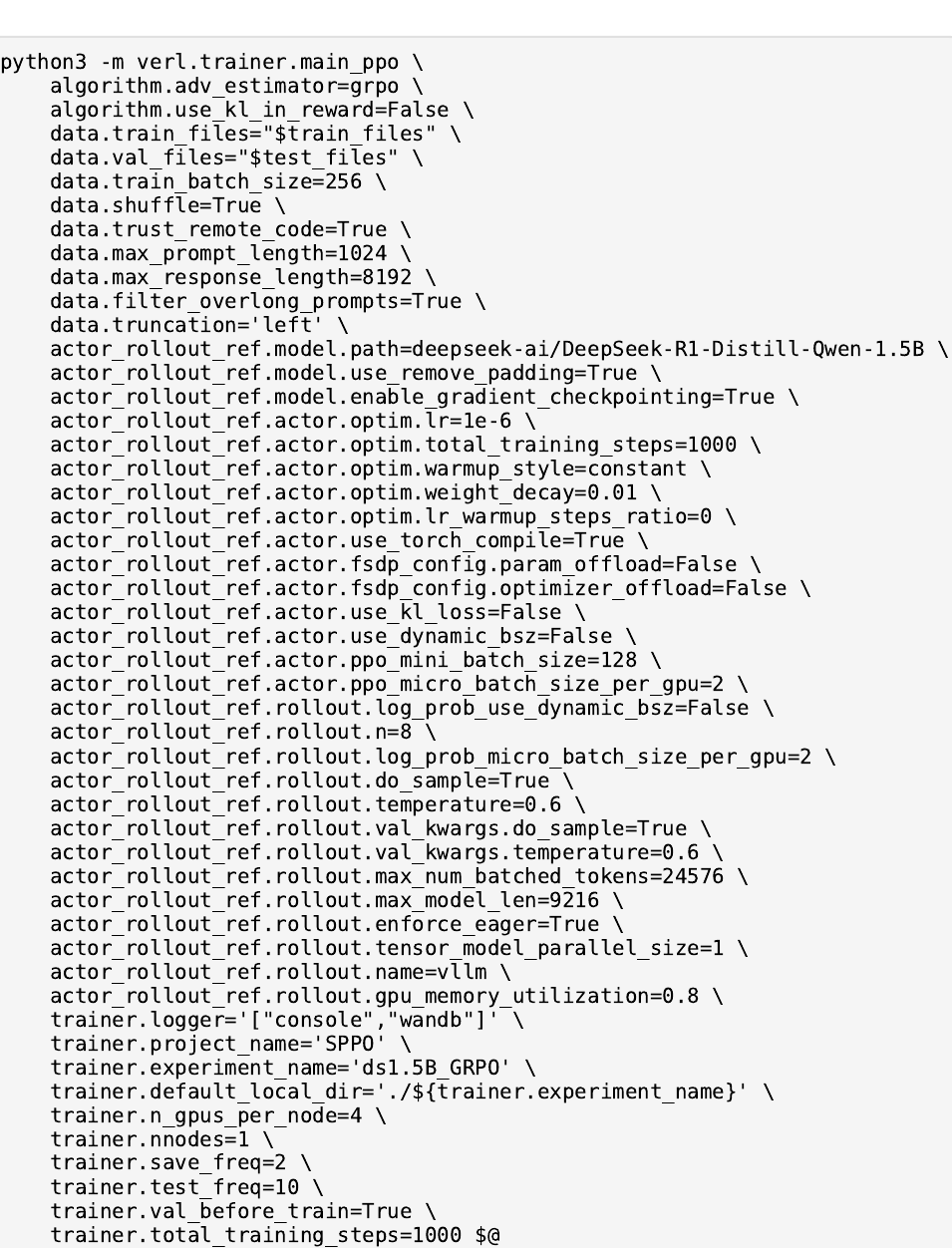}
        \captionof{figure}{Execution Command: GRPO 1.5B}
        \label{fig:cmd_grpo_1.5b}
    }
\end{minipage}

\vspace{1em}

\noindent
\begin{minipage}{\textwidth}
    {
        \centering
        \includegraphics[width=0.95\textwidth]{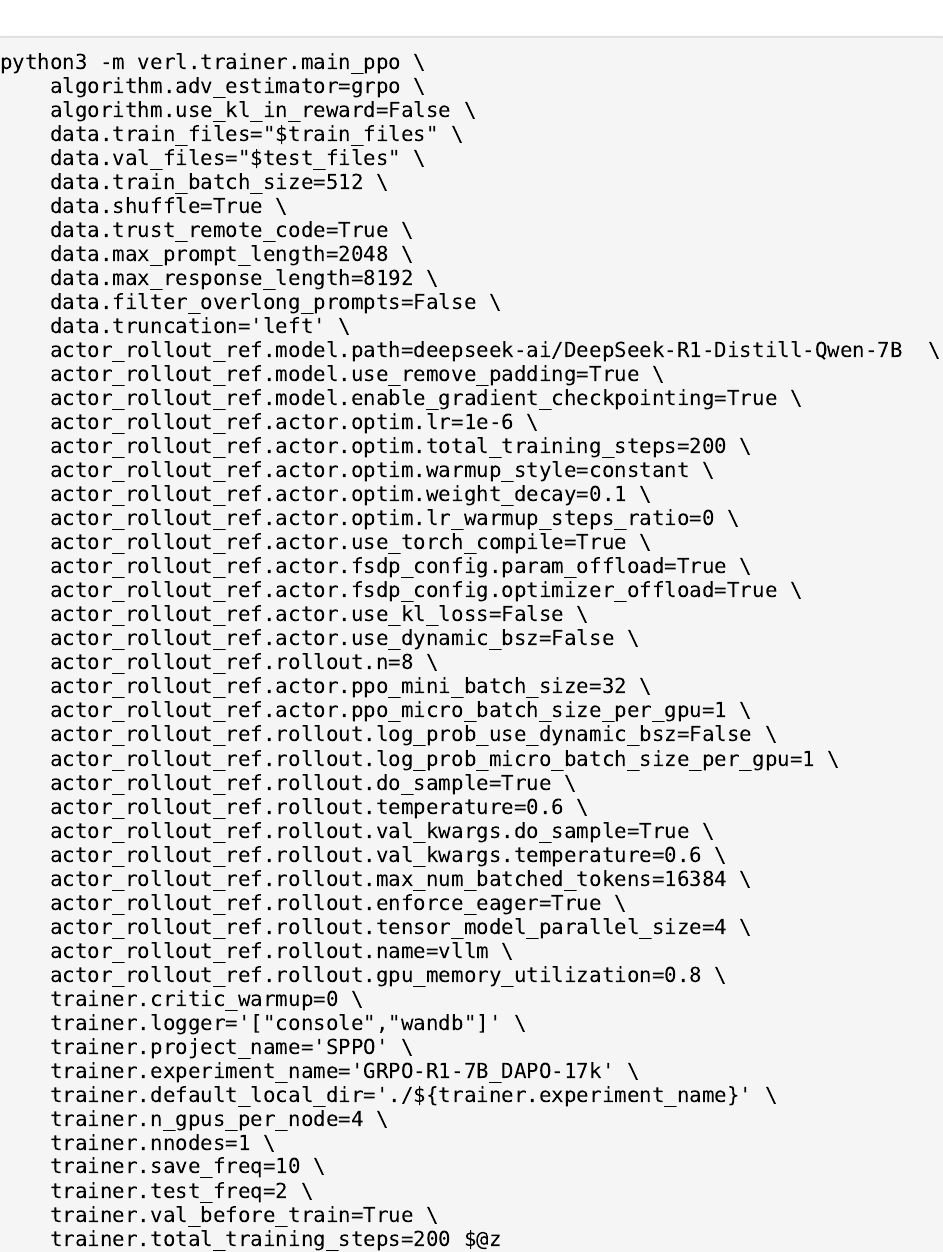}
        \captionof{figure}{Execution Command: GRPO 7B}
        \label{fig:cmd_grpo_7b}
    }
\end{minipage}

\vspace{2em}

\subsection{Standard PPO}

{
    \centering
    \includegraphics[width=0.90\textwidth]{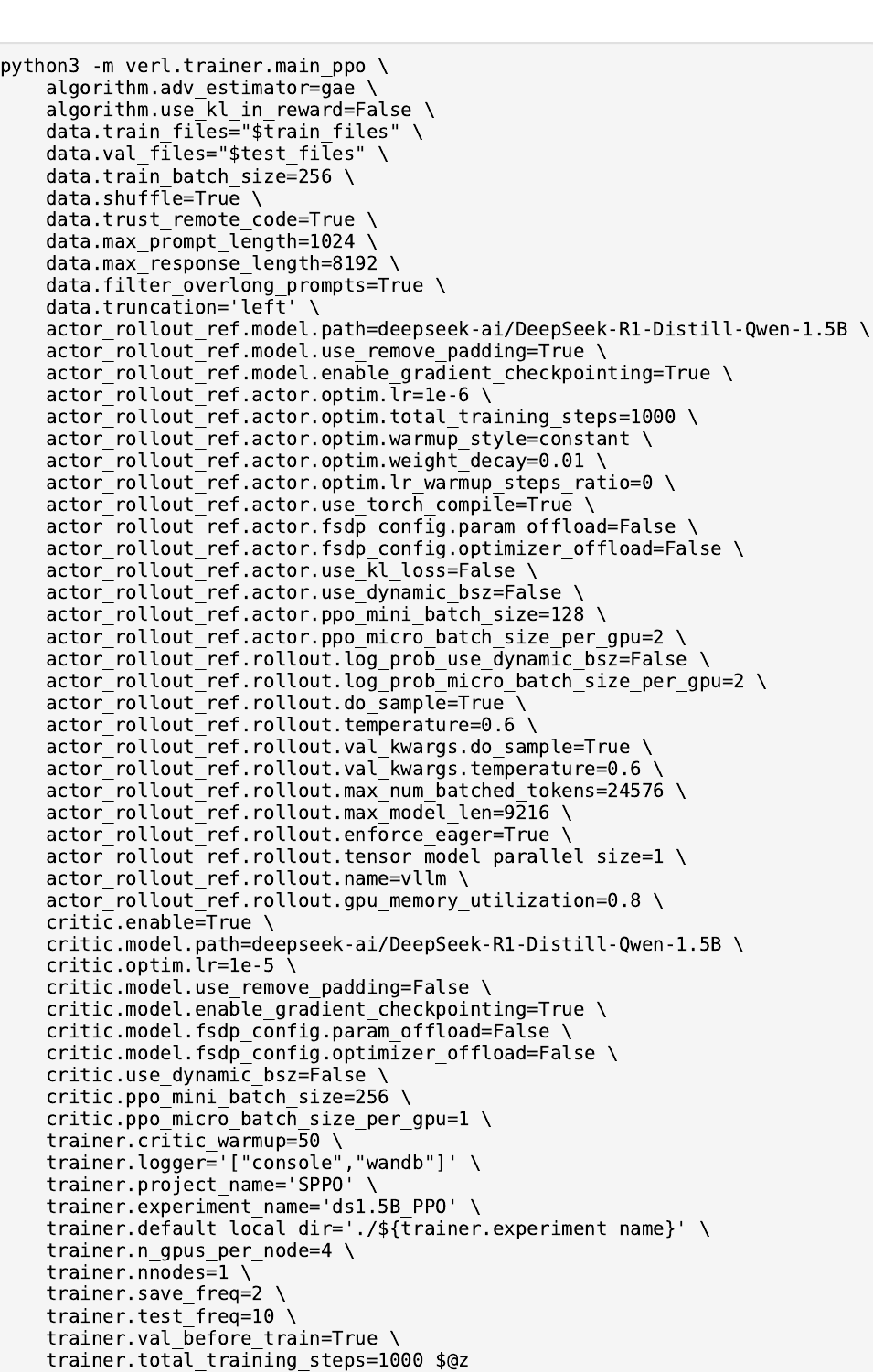}
    \captionof{figure}{Execution Command: Standard PPO 1.5B}
    \label{fig:cmd_ppo_1.5b}
}
\vspace{1em}

{
    \centering
    \includegraphics[width=0.90\textwidth]{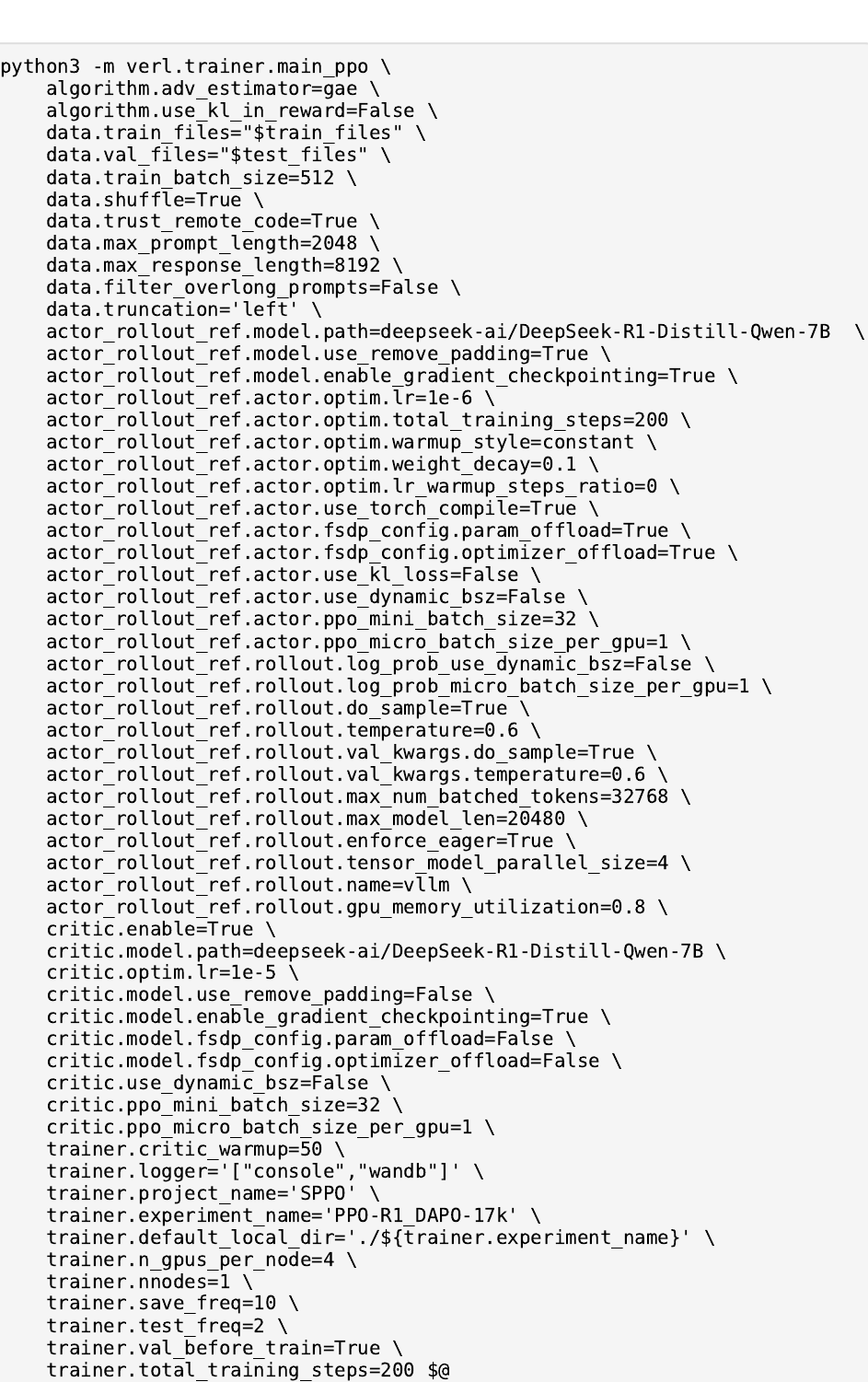}
    \captionof{figure}{Execution Command: Standard PPO 7B}
    \label{fig:cmd_ppo_7b}
}
\vspace{2em}

\noindent
\begin{minipage}{\textwidth}
    \subsection{RLOO Baselines}
    \label{app:rloo}

    \vspace{0.5em}
    
    {
        \centering
        \includegraphics[width=0.9\textwidth]{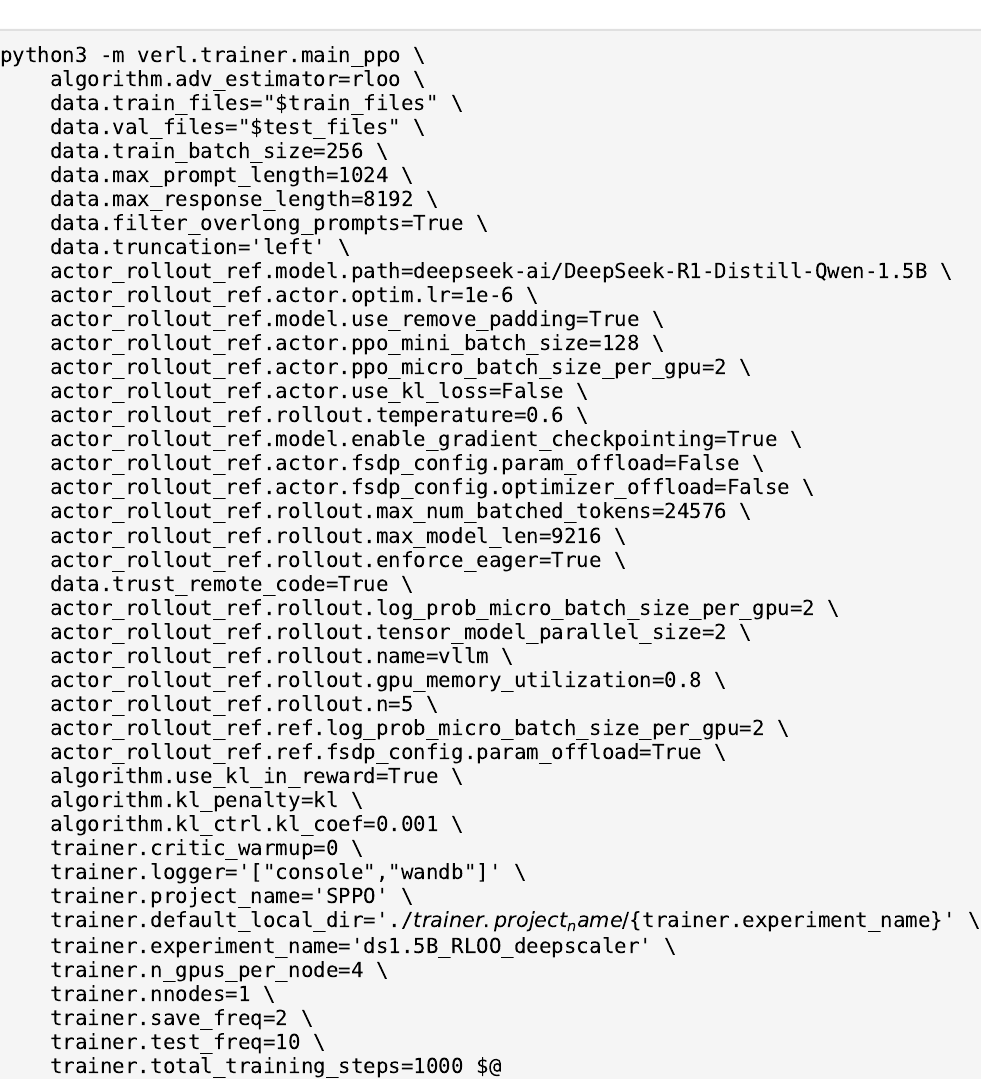}
        \captionof{figure}{Execution Command: RLOO 1.5B}
        \label{fig:cmd_rloo_1.5b}
    }
\end{minipage}

\vspace{2em}

\noindent
\begin{minipage}{\textwidth}
    {
        \centering
        \includegraphics[width=0.9\textwidth]{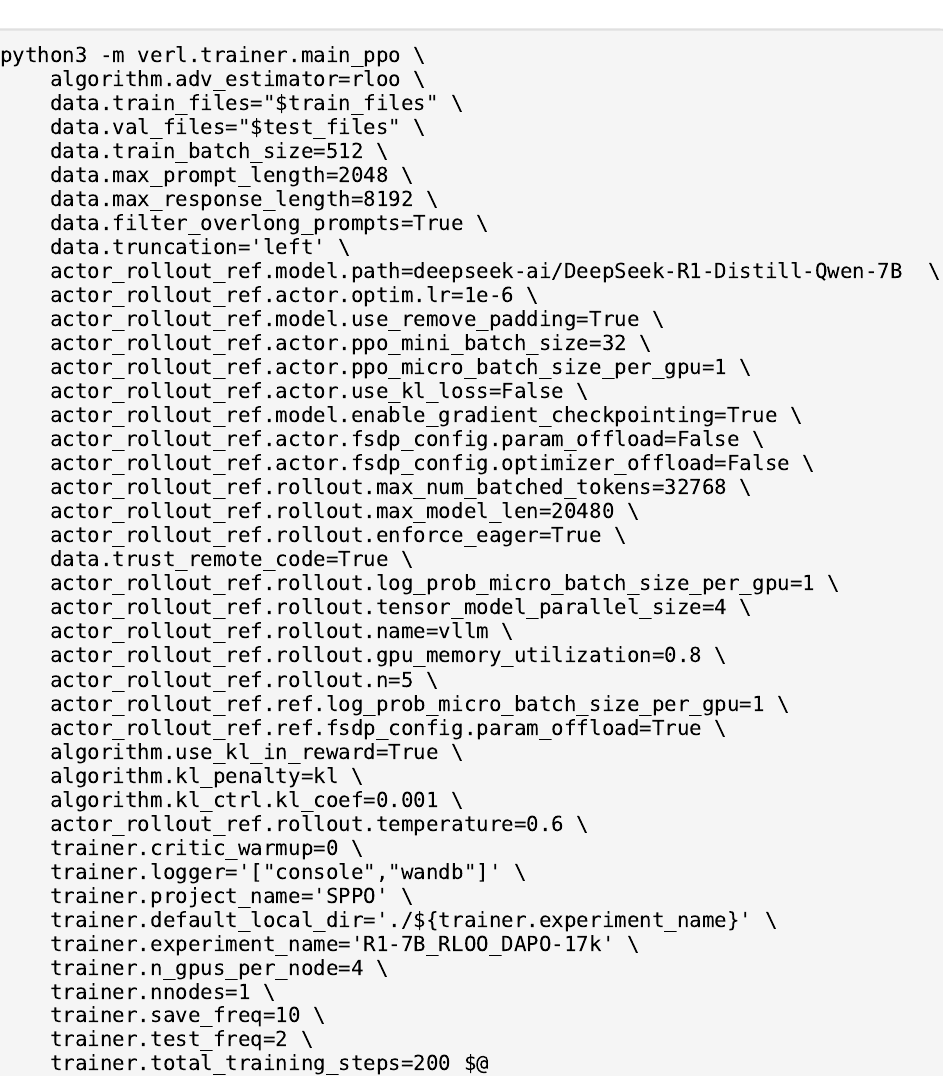}
        \captionof{figure}{Execution Command: RLOO 7B}
        \label{fig:cmd_rloo_7b}
    }
\end{minipage}

\noindent
\begin{minipage}{\textwidth}
    \subsection{ReMax Baselines}
    
    \vspace{0.5em}

    {
        \centering
        \includegraphics[width=0.95\textwidth]{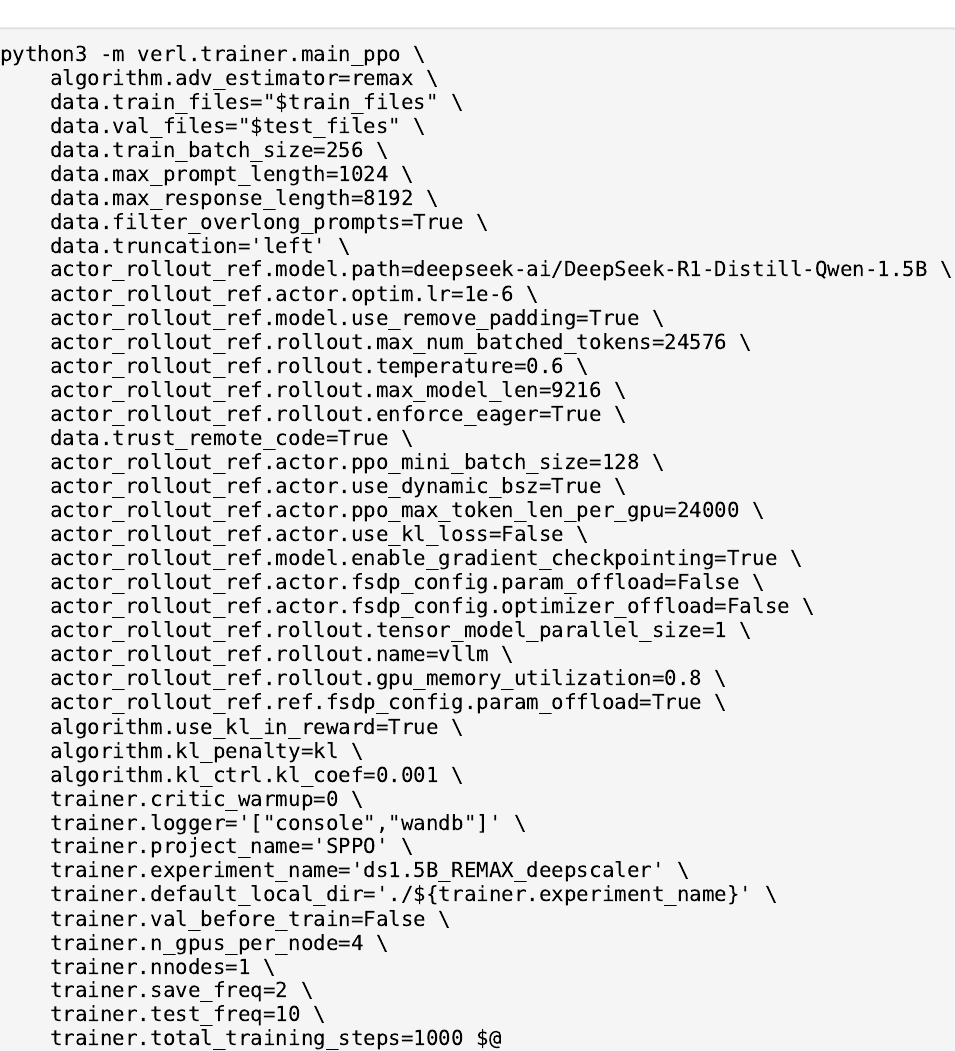}
        \captionof{figure}{Execution Command: ReMax 1.5B}
        \label{fig:cmd_remax_1.5b}
    }
\end{minipage}

\vspace{1em}

\noindent
\begin{minipage}{\textwidth}
    {
        \centering
        \includegraphics[width=0.95\textwidth]{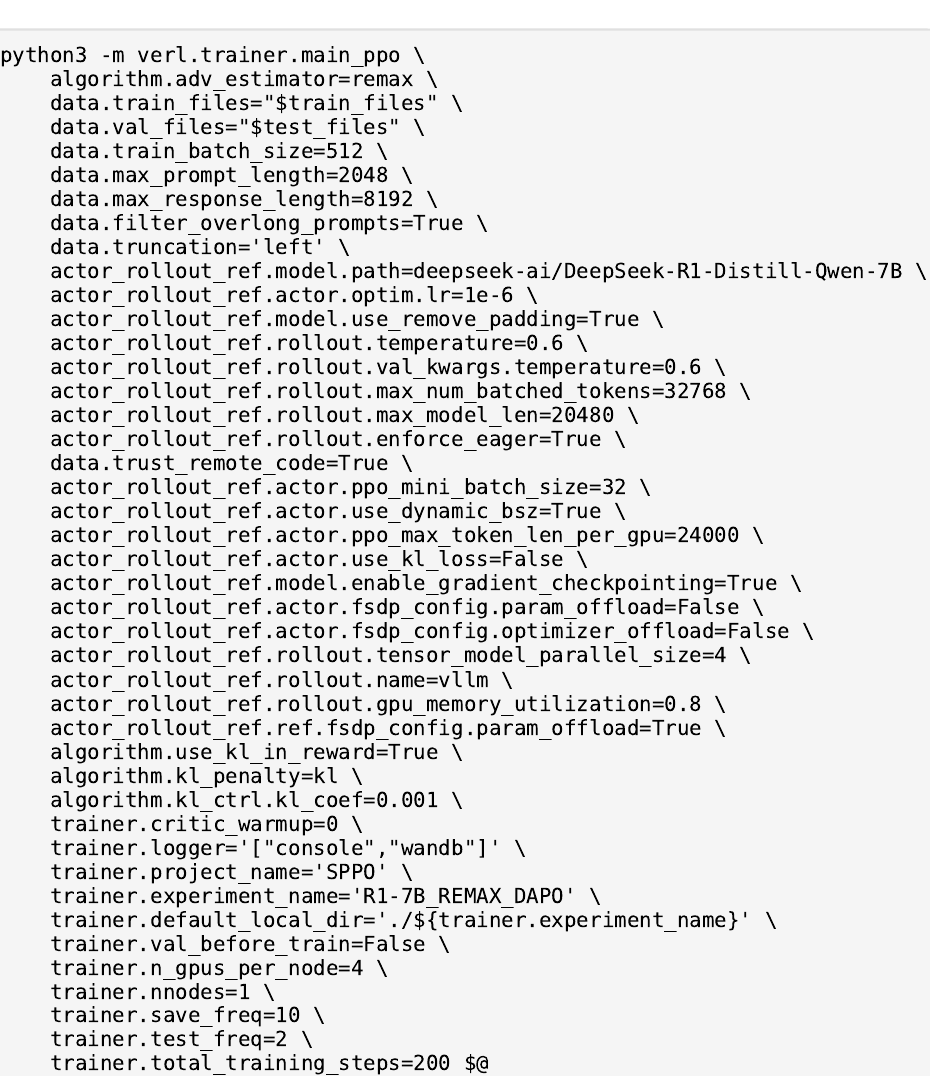}
        \captionof{figure}{Execution Command: ReMax 7B}
        \label{fig:cmd_remax_7b}
    }
\end{minipage}

\end{document}